\documentclass[11pt, letterpaper, logo]{googledeepmind}

\PassOptionsToPackage{comma,numbers,sort,compress}{natbib}

\usepackage[comma,numbers,sort,compress]{natbib}

\usepackage{hyperref}[citecolor=magenta]
\usepackage{multirow}

\hypersetup{
    colorlinks = true,
    citecolor = {magenta},
}

\pdftrailerid{redacted}

\makeatletter
\renewcommand\bibentry[1]{\nocite{#1}{\frenchspacing\@nameuse{BR@r@#1\@extra@b@citeb}}}
\makeatother

\usepackage{kantlipsum, lipsum}
\usepackage{dsfont}
\usepackage{gdm-colors}
\usepackage{subfigure}
\usepackage{bbm}
\usepackage{wrapfig}
\usepackage{colortbl}
\usepackage{xcolor}
\usepackage{graphicx}
\usepackage{caption}
\usepackage[percent]{overpic}
\usepackage{afterpage}

\usepackage[most,skins,theorems]{tcolorbox}

\tcbset{
  aibox/.style={
    width=\linewidth,
    top=8pt,
    bottom=4pt,
    colback=blue!6!white,
    colframe=black,
    colbacktitle=black,
    enhanced,
    center,
    attach boxed title to top left={yshift=-0.1in,xshift=0.15in},
    boxed title style={boxrule=0pt,colframe=white,},
  }
}
\newtcolorbox{AIbox}[2][]{aibox,title=#2,#1}
\definecolor{lightblue}{rgb}{0.22,0.45,0.70}

\usepackage{xspace}

\definecolor{WS1}{HTML}{2332F2}
\definecolor{WS2}{HTML}{344FF1}
\definecolor{WS3}{HTML}{466CF1}
\definecolor{WS4}{HTML}{578AF0}
\definecolor{WS5}{HTML}{69A7F0}
\definecolor{WS6}{HTML}{7AC4EF}
\definecolor{WS7}{HTML}{79C6D9}
\definecolor{WS8}{HTML}{78C8C4}
\definecolor{WS9}{HTML}{76CBAE}
\definecolor{WS10}{HTML}{75CD99}
\definecolor{WS11}{HTML}{74CF83}

\newcommand{\oursc}{%
  \textbf{%
    \textcolor{WS1}{W}\textcolor{WS2}{o}\textcolor{WS3}{r}\textcolor{WS4}{l}\textcolor{WS5}{d}%
    \textcolor{WS6}{S}\textcolor{WS7}{t}\textcolor{WS8}{r}\textcolor{WS9}{i}\textcolor{WS10}{n}\textcolor{WS11}{g}%
  }\xspace
}

\newcommand{\ours}{WorldString\xspace}

\graphicspath{{figures/}}

\title{Actionable World Representation}

\contributions{Kunqi Xu led the data preparation pipeline for the Articulable, Skinning, and Soft Object tasks, including model integration, and conducted all experiments except the Dr.Robot baseline. Xueyan Zou led model development and validation on articulable objects. Jitao Li conducted the Dr.Robot baseline experiments. Sifei Liu, Jianglong Ye, and Isabella Liu contributed to early idea formulation with expertise in 3D representation and robot learning. Tianshu Tang developed the theoretical formulation in the methods section.}

\author[$\vardiamondsuit$1]{Kunqi Xu}
\author[2]{Jitao Li}
\author[3]{Jianglong Ye}
\author[1]{Tianshu Tang}
\author[3]{Isabella Liu}
\author[4]{Sifei Liu}
\author[$\vardiamondsuit$1]{Xueyan Zou}

\affil[1]{Tsinghua University - IEI Lab}
\affil[2]{CalTech}
\affil[3]{UC San Diego}
\affil[4]{NVIDIA}

\affil[ ]{\protect\\[-0.2em]
$\vardiamondsuit$ Equal contribution\protect\\
\href{https://worldstring-iei.github.io/}{https://worldstring-iei.github.io/}}
\vspace{-7pt}

\begin{abstract}
Inspired by the emergent behaviors in large language models that generalized human intelligence, the research community is pursuing similar emergent capabilities within world models, with a emphasis on modeling the physical world. Within the scope of physical world model, objects are the fundamental primitives that constitute physical reality. From humans to computers, nearly everything we interact with is an object. These objects are rarely static; they are actionable entities with varying states determined by their intrinsic properties. While current methods approach object action states either via video generation or dynamic scene reconstruction, none explicitly model this basic element in a unified, principled way to build an actionable object representation. We propose \oursc, a neural architecture capable of modeling the state manifold of real-world objects by learning directly from point clouds or RGB-D video streams. Serving as a versatile digital twin, it acts as a foundational building block for physical world models; thus, we name it \ours. Sweetly, its fully differentiable structure seamlessly enables future integration with policy learning and neural dynamics.
\end{abstract}
\begin{document}

\maketitle

\begin{figure}[ht]
    \centering
    \includegraphics[width=\textwidth]{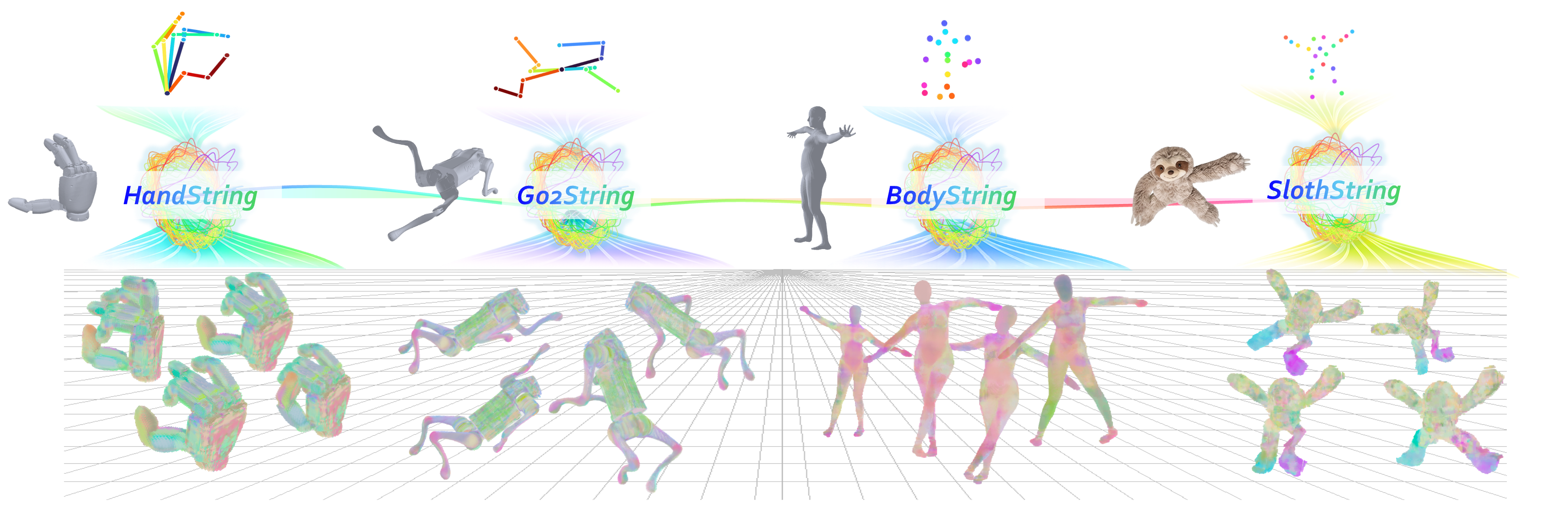}
    \caption{\textbf{Our method \oursc is a \textcolor{red}{Neural based Interactive Digital Twin} of Skinning, Articulable, and Soft objects with state as the prompt input and 3D point cloud as output.}}
    \vspace{-10pt}
\end{figure}

\section{Introduction}
\label{sec:intro}
Recent breakthroughs in large models have demonstrated strong conceptual-world modeling, but this does not automatically yield grounded physical understanding—motivating the exploration of physical world model.

\begin{figure}[h]
    \centering
    \includegraphics[width=1.\linewidth]{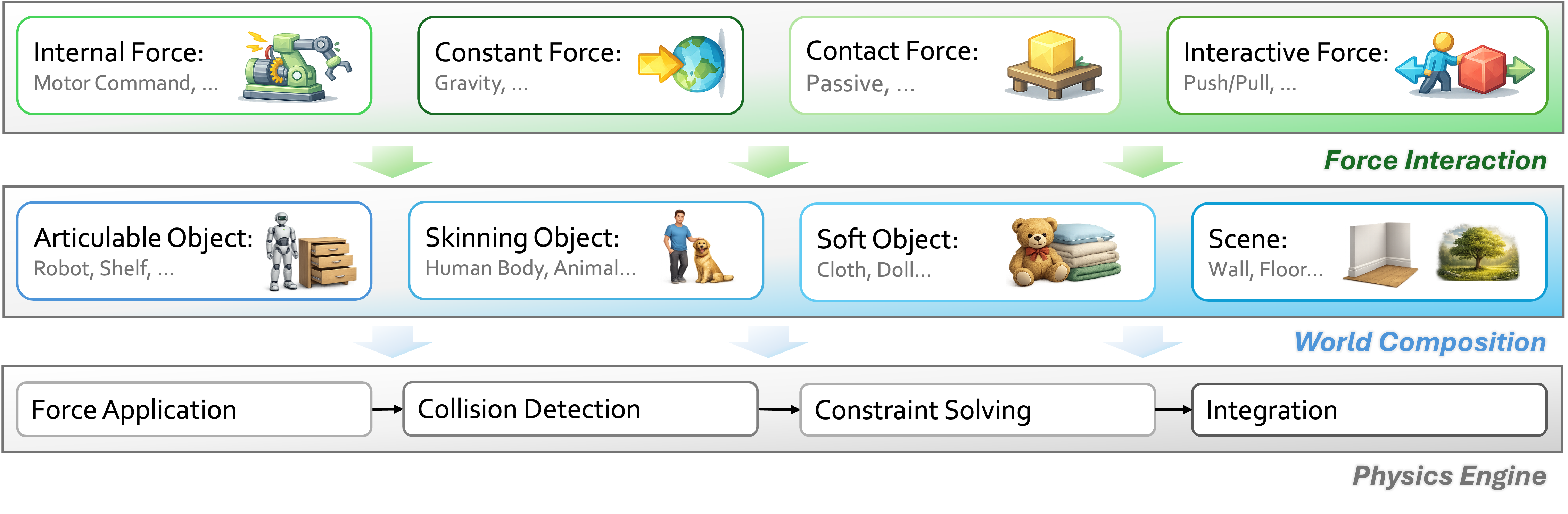}
    \vspace{-5pt}
    \caption{Position of object representation under the scope of physical world model. We position objects as the basic units of neural physical simulation. Each object representation encodes physical and semantic properties, while forces from actuation, contacts, and external interactions define how objects influence one another. A neural simulation core then aggregates forces, reasons over interactions and constraints, and rolls the world state forward over time.}
    \label{fig:concept}
\end{figure}

A physical world model serves as an agent's internal representation of its environment, capturing action-conditioned dynamics to predict future states and observations for planning, reasoning, and action~\cite{sakagami2023roboticwm,hafner2023dreamerv3,samsami2024mastering}. As illustrated in Fig.~\ref{fig:concept}, the conceptual pipeline of a physical world model fundamentally consists of: force interaction, world composition, and the underlying physics engine. Within this hierarchy, the object representation clearly state as the building blocks for physical world model.

Physical world models are commonly approached via video generation, neural 3D reconstruction, or physics simulation. Video models deliver high-fidelity, semantically rich rollouts \cite{pmlr-v235-bruce24a,huang2025vid2world} but often lack robust physical/3D consistency and controllability \cite{bansal2024videophy,zhang2025morpheus}. Reconstruction models provide 3D-consistent scene representations \cite{kerbl3Dgaussians} yet struggle with dynamic, contact-rich interactions and generalization \cite{pmlr-v270-liu25a,Zheng_2025_CVPR}. Simulation offers physically grounded interventions \cite{long2025embodied_survey,yang2024dpsi} but faces parameterization and sim-to-real gaps \cite{aljalbout2025realitygap,xu2025deal}.

Thus, we seek a representation that is controllable and action-conditioned with minimal sim-to-real gap, while retaining structured rollouts, 3D consistency, and physically grounded interventions. Because physical rollouts are ultimately driven by discrete object states and object–object interactions, we adopt an object-aligned representation as the foundational core of world model.

In this paper, we introduce \oursc, a novel actionable world representation designed as a digital twin of the physical environment. We define ``actionable'' as the inherent capacity to act, interact, and reason. Conceived as a fundamental building block of physical reality, \ours provides a unified framework capable of modeling the dynamic states of diverse entities---including articulated, skinning, and soft objects---learned directly from real-world data.

In summary, we claim the following contributions:
\vspace{-5pt}
\begin{itemize}[leftmargin=*]
    \item We introduce \ours, an actionable world representation that learns digital twins of real-world objects directly from point clouds or RGB-D video.
    \item The \ours framework provides a novel and unified pipeline that generalizes across articulated, skinning, and soft objects.
    \item Extensive quantitative and qualitative evaluations prove \ours's effectiveness in actionable object representation and the physical interpretability of its components.
\end{itemize}

\section{Related Works}
\noindent
\textbf{World Models.}
World models were introduced as learned latent simulators for prediction and control~\cite{Ha2018WorldModels}, and continue to scale to diverse domains~\cite{Hafner2025DreamerV3}. Physical world models extend this idea toward action-conditioned, simulator-like “digital twins” for physical AI~\cite{NVIDIA2025Cosmos}. Existing approaches are broadly either \emph{top-down generative}, learning to synthesize future experience or interactive worlds~\cite{Bruce2024Genie,Yu2025WonderWorld}, or \emph{bottom-up reconstructive}, inferring explicit 3D state for prediction and manipulation~\cite{Lu2025GWM}, with recent work targeting deformable digital twins from video~\cite{Jiang2025PhysTwin,Xu2026NeuSpring}. However, these methods typically model dynamics implicitly (generation) or via dense warps/primitive trajectories (reconstruction), and none explicitly capture object deformation in a correct, unified, and controllable way across articulated, skinned, and soft regimes.

\noindent
\textbf{Dynamic 3D Reconstruction.}
Neural scene reconstruction via radiance fields was popularized by NeRF~\cite{mildenhall2021nerf} and later accelerated by explicit primitives such as 3D Gaussian Splatting (3DGS)~\cite{Kerbl2023GaussianSplatting}. While both are effective for learning 3D representations from video under static-scene assumptions, real scenes are often dynamic, prompting many dynamic extensions. Dynamic NeRFs broadly fall into \emph{temporal} methods that condition on time and learn continuous deformations~\cite{Pumarola2021DNeRF,Liu2023RoDynRF} and \emph{structured-motion} methods that introduce kinematic priors or structured latents (notably for humans/articulations)~\cite{Peng2021NeuralBody,Chen2021SNARF}. Dynamic Gaussian methods similarly include temporal formulations with time-varying Gaussians or persistent tracking~\cite{Wu2024FourDGS,Luiten2024Dynamic3DGaussians} and more structured/controllable variants via editing or sparse control~\cite{Chen2024GaussianEditor,Huang2024SCGS}. Overall, these approaches typically model motion as time-/pose-conditioned warps or per-primitive trajectories from a canonical representation, rather than explicit state-transition dynamics.

\begin{wrapfigure}{r}{0.45\textwidth}
  \centering
  \includegraphics[width=0.45\textwidth]{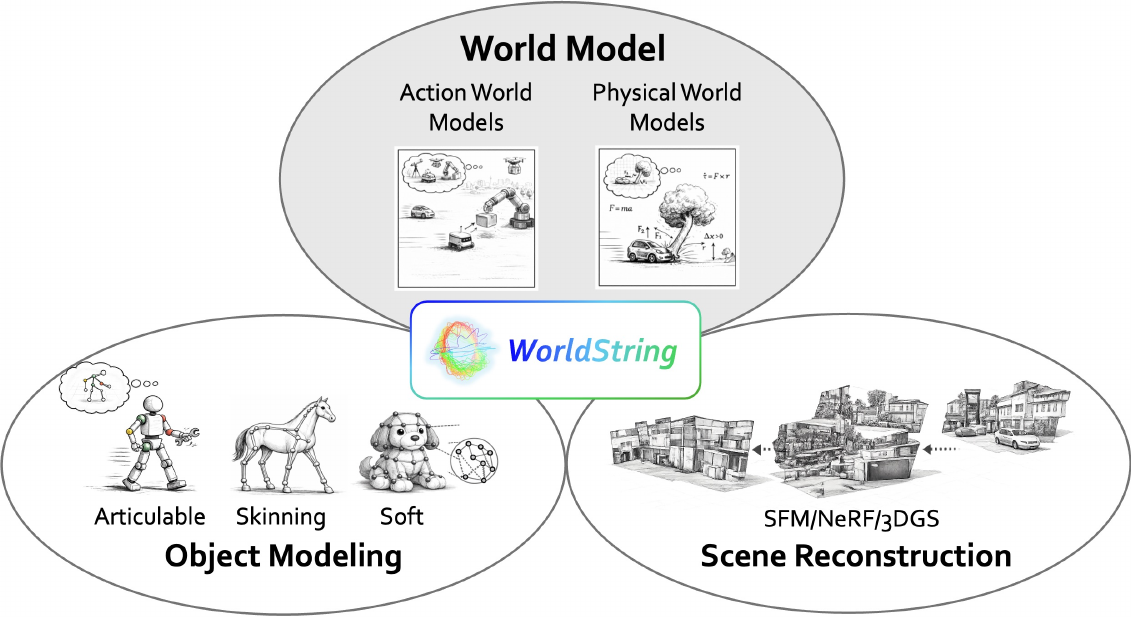}
  \caption{Position of \ours under the current related work scope.}
  \label{fig:your_label}
\end{wrapfigure}

\noindent\textbf{Classical Object Modeling.}
Classical object models range from rigid geometry to increasingly structured deformation. Static rigid shapes are represented by meshes, point clouds, voxels, or implicit fields~\cite{Hughes2014CGPP,Botsch2010PMP,GrossPfister2007PBG,Bloomenthal1997Implicit}. Articulated rigid objects are modeled as kinematic trees of links and joints (e.g., URDF), with motion parameterized by low-DoF joint configurations~\cite{LynchPark2017ModernRobotics,Spong2006RobotModelingControl,ROSURDF}. Skinned objects add a skeleton and skinning weights (e.g., LBS) to map joint motion to surface deformation~\cite{Parent2012ComputerAnimation,AkenineMoller2018RTR}. Soft/non-rigid objects exhibit high-DoF, sometimes topology-changing deformations and are traditionally handled by physics-based simulation (continuum mechanics/FEM or constraint-based dynamics), which is often costly and hard to infer from vision~\cite{Erleben2005PhysicsBasedAnimation,Bathe2014FEP,BonetWood2008NCMFEA,ZienkiewiczTaylorFox2014FEMSolid}. Recent physics-informed, video-based digital twins reconstruct deformable geometry with simulatable physical parameters for forward prediction~\cite{Jiang2025PhysTwin,Xu2026NeuSpring}. Overall, these formulations span kinematics, skinning, and physics/elasticity-based deformation, motivating learned models that bridge structure and flexibility~\cite{Parent2012ComputerAnimation,Botsch2010PMP,Erleben2005PhysicsBasedAnimation}.
\section{Method}

\begin{figure}[t]
    \centering
    \includegraphics[width=0.9\linewidth]{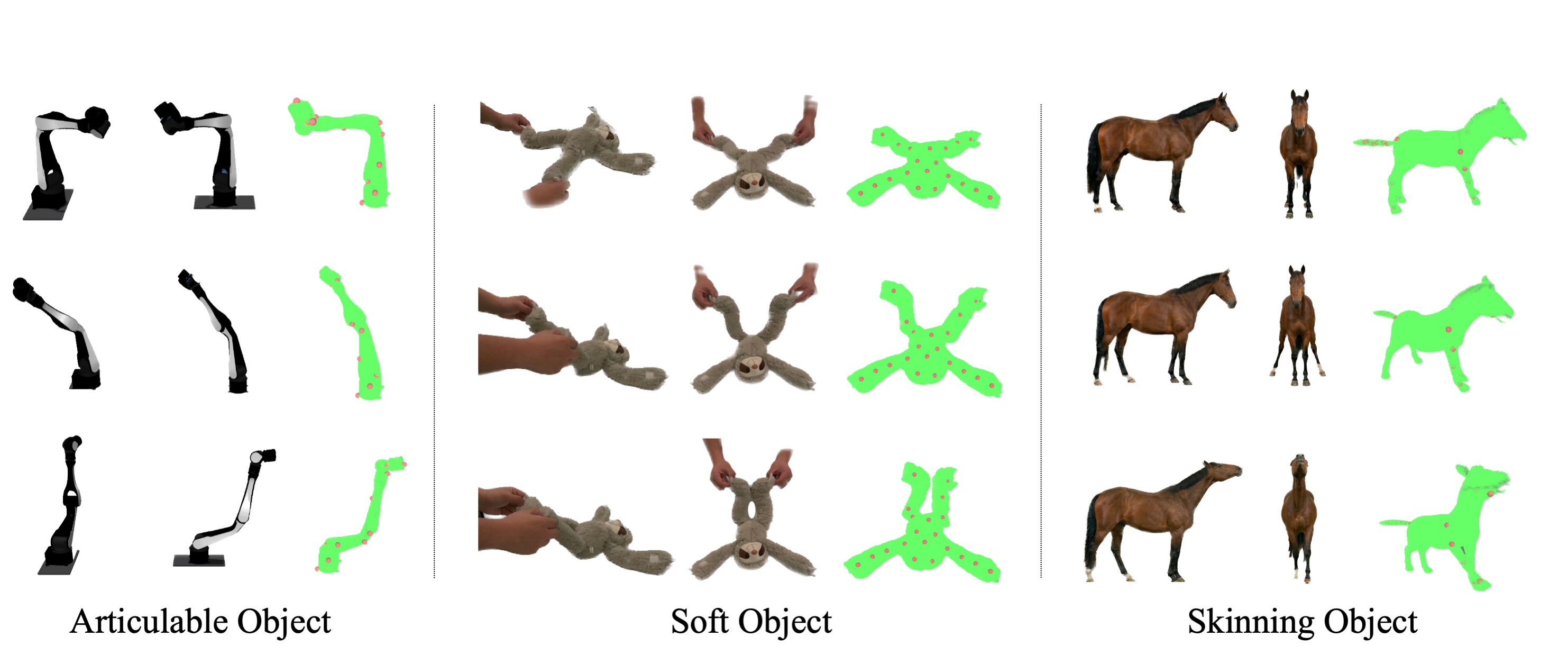}
    \caption{\textbf{Visualization of object categories.} This figure illustrates the different types of objects modeled in our framework, displaying the ground-truth point clouds and corresponding keypoints selected for each category.}
    \label{fig:object}
\end{figure}

\subsection{Background}
Under the traditional computer vision taxonomy, an image is composed of ``things'' and ``stuff.'' In the world-model narrative, a scene is instead partitioned into objects and background; typically, objects are actionable, whereas the background remains static. Formally, we can define an actionable object using the following notation: let $\Omega_* \subset\mathbb{R}^3$ denote the object's current occupancy in Cartesian space, and let $\Omega_0\subset\mathbb{R}^3$ represent its occupancy in canonical base state.
An object transition from base state to the state $u\in\mathcal U$ (e.g., joint positions) requires a deformation mapping $\Phi$, where we could formally write as:
\[
\Phi_u:\Omega_0\rightarrow \Omega_*,\qquad x=\Phi_u(y), 
\]
which sends a point $y\in\Omega_0$  in the base configuration to its world-space location $x\in\Omega_*$ under state $u$. 

In the real world, actionable objects could be summarized into three categories: \textit{Articulated Objects}, \textit{Skinned Objects}, \textit{Soft Objects}. Each of the object kind has its own state transition form as shown in Fig.~\ref{fig:object}.

\noindent\textbf{Forward Kinematics (FK).}
An articulated rigid object is a kinematic tree with joint positions $q\in\mathbb{R}^{d_q}$, i.e., $u=q$. For link $i$, let $A_i(q_i)\in SE(3)$ be the transform from its parent to $i$, and $T_j(q)=\prod_{i\in\mathcal P(j)} A_i(q_i)$ the world transform of link $j$, where $\mathcal P(j)$ is the path from root $0$ to $j$. With rest pose $q_0$ and $\Omega_0$ partitioned into link-attached subsets $\Omega_0^{(j)}$, forward kinematics yields the piecewise-rigid deformation:

\[
\Phi_u(y)=\bigl(T_j(q)T_j(q_0)^{-1}\bigr)\odot y,\qquad y\in\Omega_0^{(j)},
\]
mapping $y$ from world to link $j$'s local frame via $T_j(q_0)^{-1}$ and back via $T_j(q)$.

\noindent\textbf{Soft Object Jacobian.}
The deformation of a soft object is described by a state $u\in\mathbb{R}^{n_u}$ (e.g., nodal displacements in FEM). As $\Phi_u$ obtained from physics simulation typically has no closed form, a classical approximation is the first-order Taylor linearization around a nominal state $\bar{u}$:
\[
\Phi_{\bar{u}+\Delta u}(y)\ \approx\ \Phi_{\bar{u}}(y)\ +\ J_\Phi(y;\bar{u})\,\Delta u,
\]
where $J_\Phi(y;u)\triangleq\partial\Phi_u(y)/\partial u\in\mathbb{R}^{3\times n_u}$ is the Jacobian, measuring how the
world-space position of the material point $y$ changes linearly under an infinitesimal perturbation of the soft state.

\noindent\textbf{Linear Blend Skinning (LBS).}
A skinned object is driven by the same bone transforms $\{T_j(q)\}$ as FK, along with skinning weights $w_j:\Omega_0\to[0,1]$ satisfying $\sum_j w_j(y)=1$. 
LBS deforms a point as the weighted sum of its rigidly transformed positions under each bone:

\[
\Phi_u(y)=\sum_{j} w_j(y)\,\bigl(T_j(q)\,T_j(q_0)^{-1}\bigr)\odot y,\qquad y\in\Omega_0.
\vspace{-8pt}
\]

\begin{figure}[t]
    \centering
    \includegraphics[width=1.\linewidth]{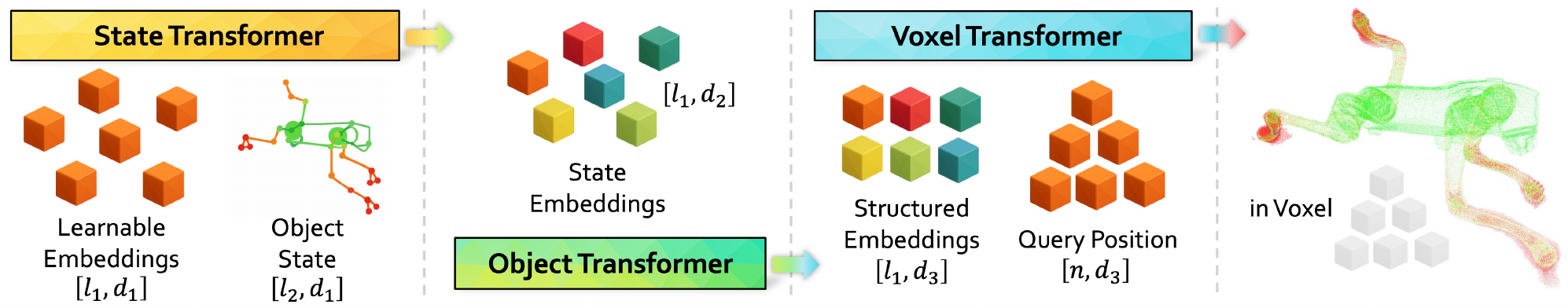}
    \caption{\textbf{\ours model pipeline.} Our fully differentiable architecture learns an actionable world representation by optimizing canonical embeddings and cascaded transformers to reconstruct the target object state.}
    \label{fig:worldstring}
    \vspace{-15pt}
\end{figure}

\subsection{Formulation}
To model actionable objects from 3D or RGB-D data, we translate the physical formulation into a fully differentiable architecture: the canonical base state $\Omega_0$ is parameterized as learnable embeddings $\omega_0 \in \mathbb{R}^{l_1 \times d_1}$ ($l$ is embedding number, and $d$ is embedding dimension), the dynamic state $u$ as sparse structural keypoints $K \in \mathbb{R}^{l_2 \times d_1}$, and the deformation mapping $\Phi_u$ as learnable transformer layers $\Phi$.

The deformation logic $\Phi$ is factorized into a two-stage transformer architecture. First, the State Transformer $\Phi_s$ utilizes cross-attention to condition the canonical base embeddings $\omega_0$ on the dynamic keypoint state $K$, computing the intermediate state embeddings $Z_s \in \mathbb{R}^{l_1 \times d_2}$:
$Z_s = \Phi_s(\omega_0, K).$
This operation injects localized keypoint constraints, effectively grounding the canonical geometry in the current pose. 

Subsequently, to propagate these localized deformations and enforce global structural coherence across the object manifold, the Object Transformer $\Phi_o$ applies self-attention over $Z_s$:
$Z_{\text{obj}} = \Phi_o(Z_s),$
yielding the structured embeddings $Z_{\text{obj}} \in \mathbb{R}^{l_1 \times d_3}$, which comprehensively encapsulate the fully deformed object within the latent space.

While the structured embeddings $Z_{\text{obj}}$ implicitly capture the deformed state, they reside in an uninterpretable latent space. To recover the explicit object geometry in Cartesian space $\Omega_*$, we employ the Voxel Transformer $\Phi_v$. We construct spatial queries $Q(x)$ from continuous 3D coordinates $x \in \mathbb{R}^3$ via positional encoding. The Voxel Transformer cross-attends these spatial queries with $Z_{\text{obj}}$ to predict the continuous occupancy field:
$O(x) = \Phi_v(Q(x), Z_{\text{obj}}),$
where $O(x) \in [0, 1]$ represents the probability that the point $x$ belongs to the object. By densely querying the workspace, we can extract the explicit voxel grid of the deformed object.

During training, we randomly sample a set of spatial points $x_i \in \mathbb{R}^3$ within the workspace, whereas during evaluation, we exhaustively query a dense voxel grid to reconstruct the complete object geometry. The framework is optimized end-to-end using a Binary Cross-Entropy (BCE) loss. Through this continuous occupancy prediction, we complete the fully differentiable pipeline, successfully mapping the implicit canonical base state $\Omega_0$ and sparse keypoints to the explicitly rendered target state $\Omega_*$.

\subsection{Generalization}
In the following paragraphs, we demonstrate that the proposed \ours model serves as a unified generalization of Forward Kinematics (FK), Linear Blend Skinning (LBS), and soft object Jacobians.

\noindent
\textbf{Sufficiency of Keypoints for Geometry Recovery}
We attach $K$ keypoints to the canonical object at locations $\{\xi_i\}_{i=1}^K\subset\Omega_0$ and observe their world positions $\Phi_u(\xi_i)\in\mathbb{R}^3$ under state $u$.

For FK and LBS, $\Phi_u$ is determined by per-link/bone rigid transforms, which are uniquely identified from at least 3 non-collinear keypoints per link/bone.

For soft objects, let $d_u(y)=\Phi_u(y)-y$ be the displacement field, assumed $L$-Lipschitz: $\|d_u(y)-d_u(y')\|\le L\|y-y'\|$ for all $y,y'\in\Omega_0$. If $\{\xi_i\}_{i=1}^K$ form a $\delta$-net of $\Omega_0$ (every $y$ is within distance $\delta$ of some $\xi_i$), then nearest-keypoint approximation $\tilde d_u(y)=d_u(\xi_{i(y)})$ satisfies $\sup_{y\in\Omega_0}\|d_u(y)-\tilde d_u(y)\|\le L\delta.$
Hence, keypoints determine the soft deformation up to an $O(L\delta)$ approximation error.

\noindent
\textbf{A unified operator view and attention as its relaxation}

Articulated, skinned, and soft objects share a unified displacement form: a convex combination of keypoint-induced updates. For any point $y\in\Omega_0$,
\vspace{-5pt}
\[
\Phi_u(y) = y + \sum_{i=1}^K \alpha_i(y;u)\,v_i(y;u),\quad
\alpha_i(y;u)\ge 0,\;\sum_{i=1}^K \alpha_i(y;u)=1,
\label{eq:unified-residual}
\vspace{-5pt}
\]
where $v_i(y;u)\in\mathbb{R}^3$ is the displacement contribution from keypoint $i$.
FK uses one-hot $\alpha_i$ selecting the owning link, and LBS uses fixed $\alpha_i=w_i(y)$.
For soft objects, while the Jacobian increment $J_\Phi(y;\bar u)\Delta u$ is not convex in general, keypoint sufficiency motivates convex interpolation of the displacement field from keypoint displacements (e.g., FEM shape functions),
$d_u(y)\approx \sum_{i=1}^K \alpha_i(y)\,d_u(\xi_i)$ with $\alpha_i(y)\ge 0$ and $\sum_i\alpha_i(y)=1$, which fits \eqref{eq:unified-residual} with $v_i(y;u)\equiv d_u(\xi_i)$.

Cross-attention is a relaxation of \eqref{eq:unified-residual}: it keeps convex mixing but replaces analytic $(\alpha_i,v_i)$ by learned, state-dependent ones.
With $q(y)$ and $\{k_i(u),\tilde v_i(y;u)\}$,
\vspace{-5pt}
\[
\mathrm{Attn}(y;u)=\sum_{i=1}^K \tilde \alpha_i(y;u)\,\tilde v_i(y;u),
\qquad
\tilde \alpha_i(y;u)=\mathrm{softmax}_i\!\big(\langle q(y),k_i(u)\rangle\big),
\label{eq:attn-mixture}
\vspace{-5pt}
\]
With the residual connection, attention naturally implements the additive form $\Phi_u(y)=y+\Delta(y)$.

\begin{figure}[t]
    \centering
    \includegraphics[width=1.\linewidth]{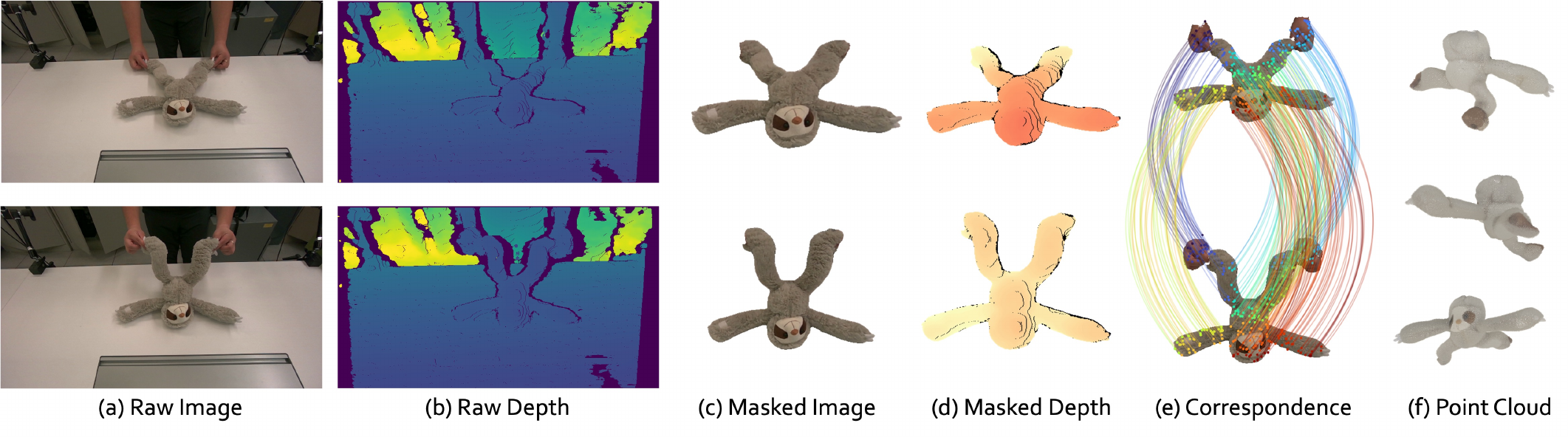}
    \caption{\textbf{\ours model learning from RGB-D video data.} The figure shows the processed data from PhysTwin \cite{Jiang2025PhysTwin}, including the raw video frames, depth maps, and masked object correspondences.}
    \label{fig:videodata}
\end{figure}

\subsection{Application: Real-World Data Acquisition}
\label{sec:data_acquisition}

To ground the differentiable representation in reality, we develop a pipeline that maps raw multi-view RGB-D observations $\mathcal{O} = \{I_t, D_t\}_{t=0}^T$, where $I_t$ and $D_t$ denote the RGB images and depth maps at frame $t$, to a sequence of paired volumetric states and keypoints $\mathcal{S} = \{(\mathcal{V}_t, \mathcal{K}_t)\}_{t=0}^T$.

\noindent\textbf{Dense 3D Tracking.} 
Following PhysTwin \cite{jiang2025phystwinphysicsinformedreconstructionsimulation}, we segment the object using Grounded-SAM2\cite{ren2024grounded} and track dense pixels via CoTracker\cite{karaev24cotracker3}. By unprojecting these 2D trajectories into 3D using the depth maps $D_t$ and camera intrinsics, we obtain a temporal sequence of dense 3D point clouds $\mathcal{P}_t = \{\mathbf{p}_{i,t} \in \mathbb{R}^3\}_{i=1}^N$. Here, $i$ denotes the identity index of a consistently tracked point across all frames, ensuring temporal correspondence.

\noindent\textbf{Geometric Initialization and Anchoring.} 
For the initial frame $t=0$, a canonical mesh $\mathcal{M}_0$ is generated via TRELLIS\cite{xiang2024structured} and refined to fit $\mathcal{P}_0$ through coarse-to-fine registration. We define the structural anchors by selecting a sparse set of keypoints $\mathcal{K}_0 \subset \mathcal{P}_0$ via Farthest Point Sampling (FPS). These keypoints $\mathcal{K}_t$ are naturally propagated through time following the tracked displacements in $\mathcal{P}_t$, ensuring a fixed relative topology on the object manifold.

\noindent\textbf{Vertex Warping and Voxelization.} 
The sequence of dense volumetric targets $\mathcal{V}_t$ is generated by warping the canonical mesh $\mathcal{M}_0$ to each frame $t$. For each vertex $\mathbf{v} \in \mathcal{M}_0$, its position at time $t$ is computed via displacement interpolation:
\vspace{-5pt}
\[
\mathbf{v}_t = \mathbf{v}_0 + \sum_{j \in \mathcal{N}(\mathbf{v})} w_j (\mathbf{p}_{j,t} - \mathbf{p}_{j,0})
\vspace{-5pt}
\]
where $\mathcal{N}(\mathbf{v})$ denotes indexs of the $k$-nearest tracking points in $\mathcal{P}_0$ for $\mathbf{v}$, and $w_j$ are skinning weights derived from inverse-distance weighting. The warped mesh $\mathcal{M}_t$ is then voxelized to form the occupancy target $\mathcal{V}_t \in \{0, 1\}^{R^3}$.

\noindent\textbf{Cross-Sequence Alignment.} 
To aggregate diverse videos, we enforce cross-sequence consistency of $\mathcal{K}$ using RoMa\cite{edstedt2024roma}. By establishing pixel correspondences between initial frames of different sequences, we anchor a unified keypoint set across the entire dataset, enabling the AWR model to learn from various interaction trajectories within a consistent structural coordinate system.
\section{Experiments}

\subsection{Reconstruction of Complex 3D Rigid Shapes}

\label{sec:rigid_reconstruction}
To evaluate \ours's fundamental geometric modeling capacity, we first assess the reconstruction of complex rigid objects, including the \textit{Utah Teapot}, \textit{Stanford Bunny}, \textit{Armadillo}, and \textit{Lucy}~\cite{10.1145/360349.360353,10.1145/192161.192241,10.1145/237170.237270,10.1145/344779.344849}. While this setup involves only a single pose, it serves as a rigorous test for fitting intricate topologies. As visualized in Table ~\ref{table:rigid_object}, our model accurately captures the global manifold and distinctive features of these benchmarks. In the error gradient maps, \textbf{blue} regions indicate near-perfect alignment with the ground truth, while \textbf{pink} highlights localized spatial deviations. The results demonstrate that \ours recovers the overall structure with high fidelity, with minor discrepancies appearing only in extremely fine-grained crevices and high-curvature furrows. This provides a solid geometric foundation for the subsequent experiments.

\begin{table}[h]
\centering
\caption{Results of \ours rigid shape reconstruction.}
\renewcommand{\arraystretch}{1.2}
\setlength{\tabcolsep}{4pt}
\resizebox{\linewidth}{!}{
\begin{tabular}{l|cccc|cccc|cccc|cccc}
\toprule
& \multicolumn{4}{c|}{\underline{\textbf{Utah Teapot}}}
& \multicolumn{4}{c|}{\underline{\textbf{Stanford Bunny}}}
& \multicolumn{4}{c|}{\underline{\textbf{Armadillo}}}
& \multicolumn{4}{c}{\underline{\textbf{Lucy}}} \\

\textbf{Object}
& \multicolumn{4}{c|}{Easy}
& \multicolumn{4}{c|}{Medium}
& \multicolumn{4}{c|}{Hard}
& \multicolumn{4}{c}{Hard} \\
\hline
\textbf{Ground Truth}
& \multicolumn{4}{c|}{\includegraphics[width=0.25\textwidth]{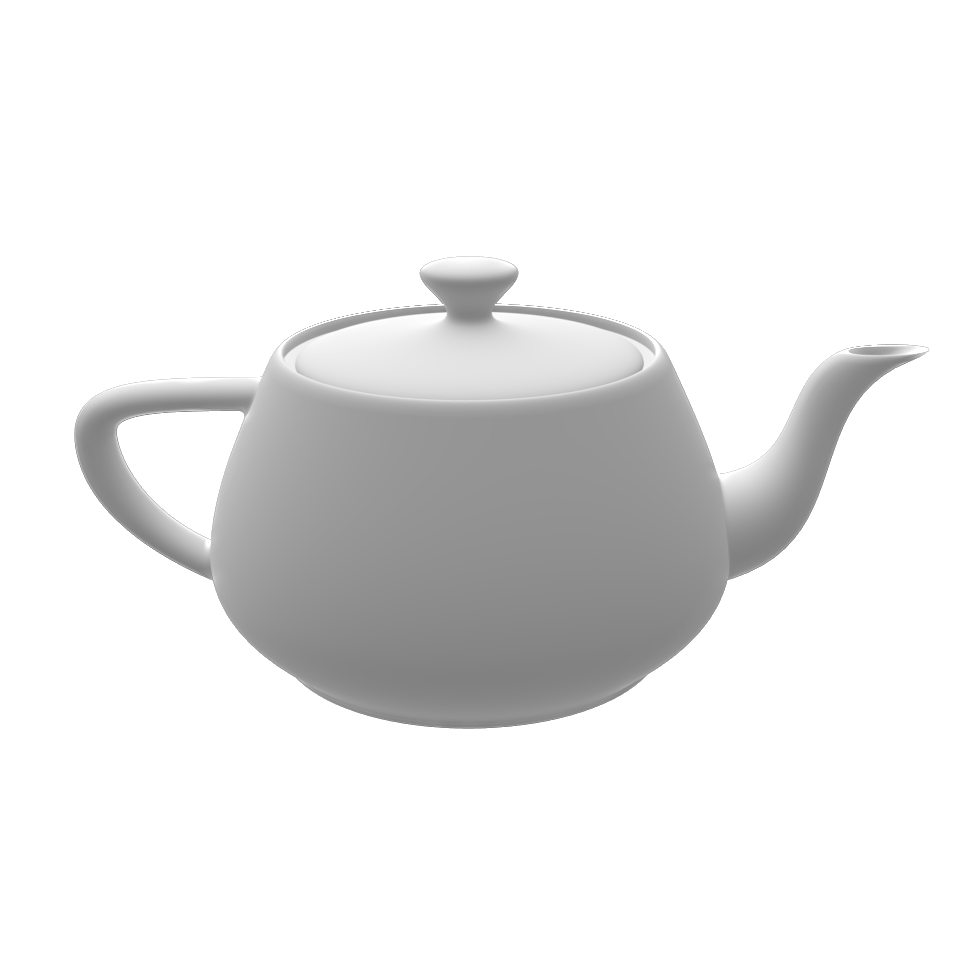}}
& \multicolumn{4}{c|}{\includegraphics[width=0.25\textwidth]{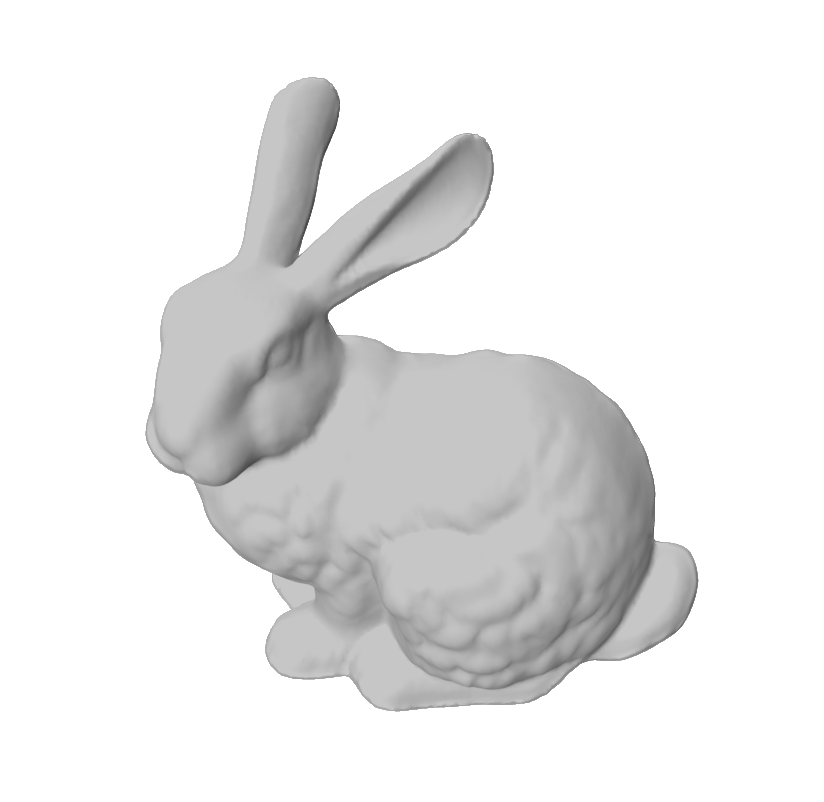}}
& \multicolumn{4}{c|}{\includegraphics[width=0.25\textwidth]{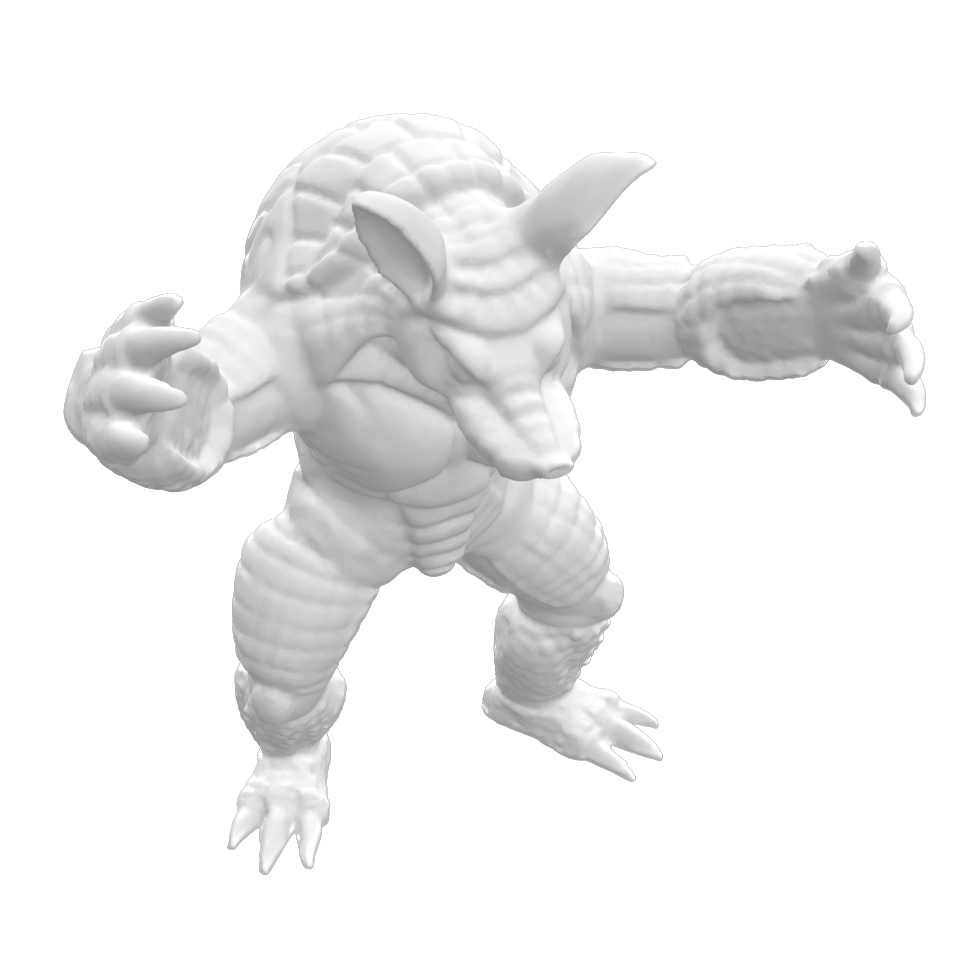}}
& \multicolumn{4}{c}{\includegraphics[width=0.25\textwidth]{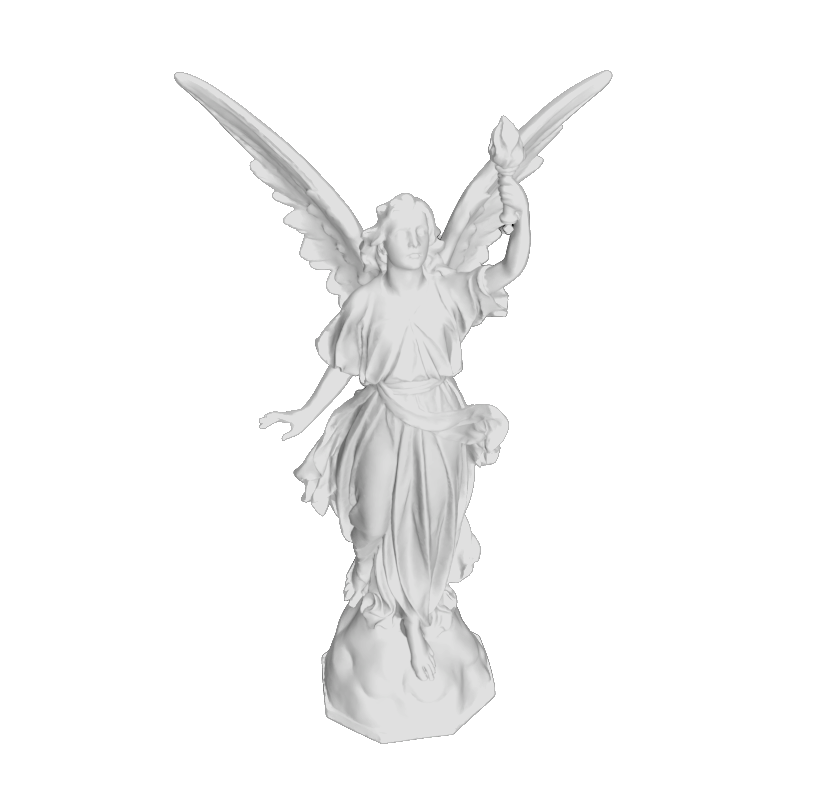}} \\
\hline
\textbf{Qualitative}
& \multicolumn{4}{c|}{\includegraphics[width=0.25\textwidth]{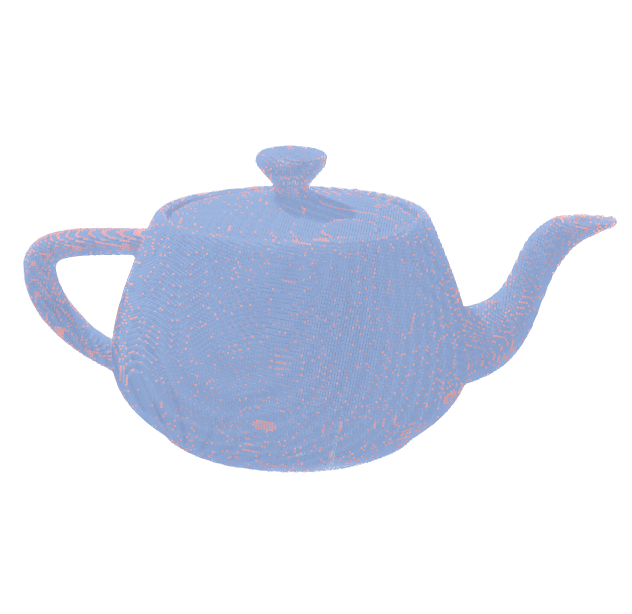}}
& \multicolumn{4}{c|}{\includegraphics[width=0.25\textwidth]{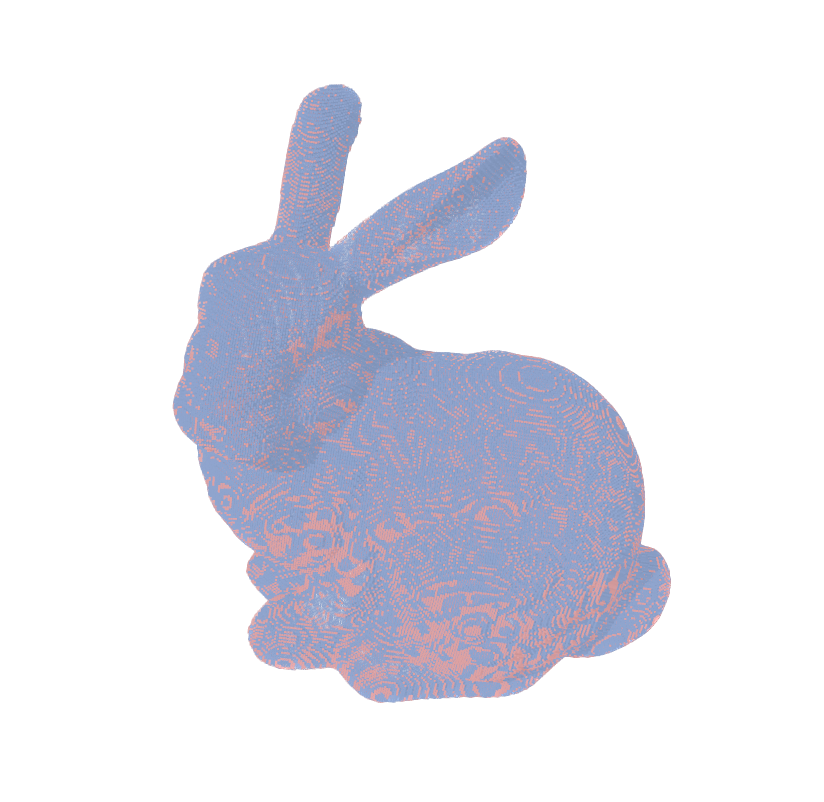}}
& \multicolumn{4}{c|}{\includegraphics[width=0.25\textwidth]{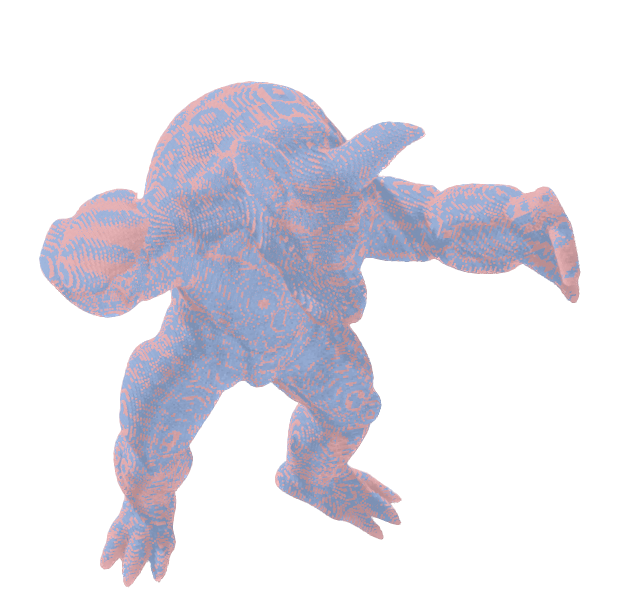}}
& \multicolumn{4}{c}{\includegraphics[width=0.25\textwidth]{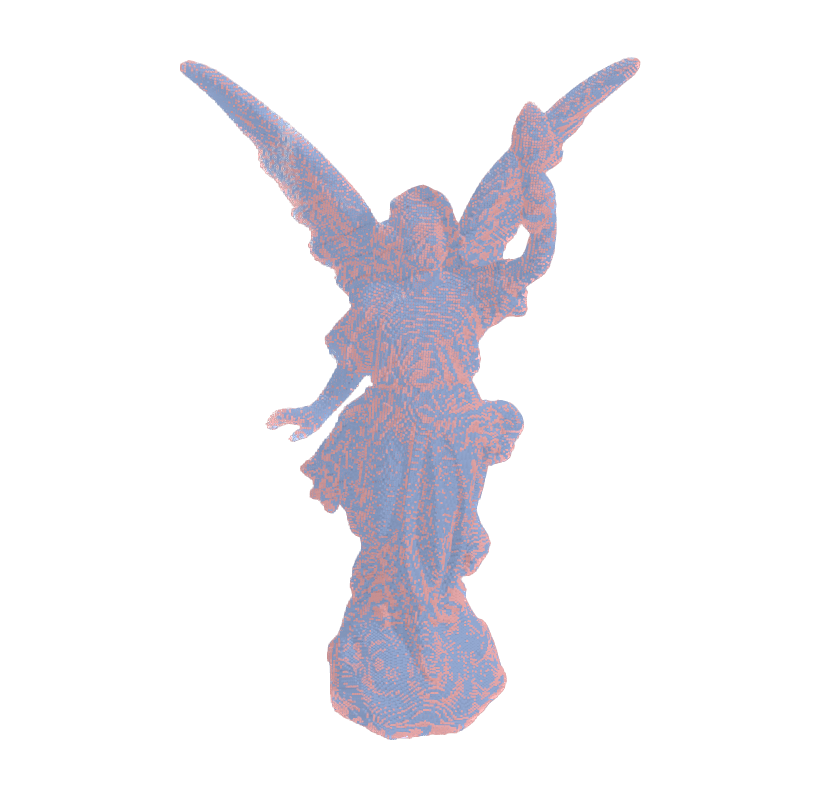}} \\
\hline
\multirow{2}{*}{\textbf{Quantitative}}
& $\mathrm{IoU}\!\uparrow$ & $F_{1}\!\uparrow$ & $\mathcal{P}\!\uparrow$ & $\mathcal{R}\!\uparrow$
& $\mathrm{IoU}\!\uparrow$ & $F_{1}\!\uparrow$ & $\mathcal{P}\!\uparrow$ & $\mathcal{R}\!\uparrow$
& $\mathrm{IoU}\!\uparrow$ & $F_{1}\!\uparrow$ & $\mathcal{P}\!\uparrow$ & $\mathcal{R}\!\uparrow$
& $\mathrm{IoU}\!\uparrow$ & $F_{1}\!\uparrow$ & $\mathcal{P}\!\uparrow$ & $\mathcal{R}\!\uparrow$ \\
& 92.17 & 95.92 & 92.66 & 99.42
& 75.38 & 85.96 & 75.87 & 99.15
& 67.36 & 80.50 & 67.76 & 99.12
& 70.20 & 82.49 & 70.74 & 98.93 \\
\bottomrule
\end{tabular}
}
\label{table:rigid_object}
\end{table}

\subsection{Baselines}
In baseline selection, we implement two retrieval-based baselines for all kinds of objects, Dr. Robot for Articulated objects, NSDP for Skinning-based humans and animals, and HALO for human hand:
\begin{itemize}[leftmargin=15pt]
    \item \textbf{Nearest Neighbor (NN):} We compress the training set by clustering the keypoint trajectories into $K$ centroids using the K-means algorithm. For each centroid, the training frame closest to the cluster center is stored. The total disk space occupied by the stored states in the baselines is restricted to not exceed the size of our trained WorldString model weights. At test time, given a new keypoint input, the model retrieves the shape point cloud from the stored state that has the most similar keypoint configuration.
    \item \textbf{Optimized NN (Optim. NN):} Building upon the NN baseline, this approach further refines the retrieved shape to accommodate unseen poses. After identifying the nearest stored state, we apply Inverse Distance Weighting(IDW) to interpolate the deformation field across the entire shape.
    \item \textbf{Dr.~Robot}~\cite{liu2024differentiablerobotrendering}: A differentiable articulated robot renderer that represents appearance with 3D Gaussian splatting in a canonical configuration and deforms it with kinematics-aware linear blend skinning and differentiable forward kinematics. We use it for articulated rigid objects.
    \item \textbf{NSDP}~\cite{tang2022neural}: Neural Shape Deformation Priors predicts mesh deformations from sparse user handles by learning a composition of local surface deformations with transformer-based deformation networks and latent codes anchored in 3D space. We use it as a learned deformation prior for skinning-based humans and animals.
    \item \textbf{HALO}~\cite{karunratanakul2021halo}: A skeleton-driven neural occupancy model that maps 3D hand joint locations to an implicit surface of the posed hand, enabling dense geometry from skeletal input alone. We adopt it for human hand experiments.
\end{itemize}

\subsection{Articulated Objects and Robots}
\noindent
In this section, we verify the how WorldString performs on articulated objects(Xhand, Airbot Play and two IKEA Cabinets). As summarized in Table~\ref{table:articulated_object}, WorldString consistently outperforms both retrieval-based baselines across various articulated categories. WorldString's continuous neural field effectively captures the piecewise rigid kinematics of articulated joints. The high IoU and F1-scores indicate that our model maintains the structural integrity of rigid parts during rotation and translation, providing a more coherent representation of joint limits and connectivity compared to baselines.

\begin{figure}
    \centering
    \includegraphics[width=0.9\linewidth]{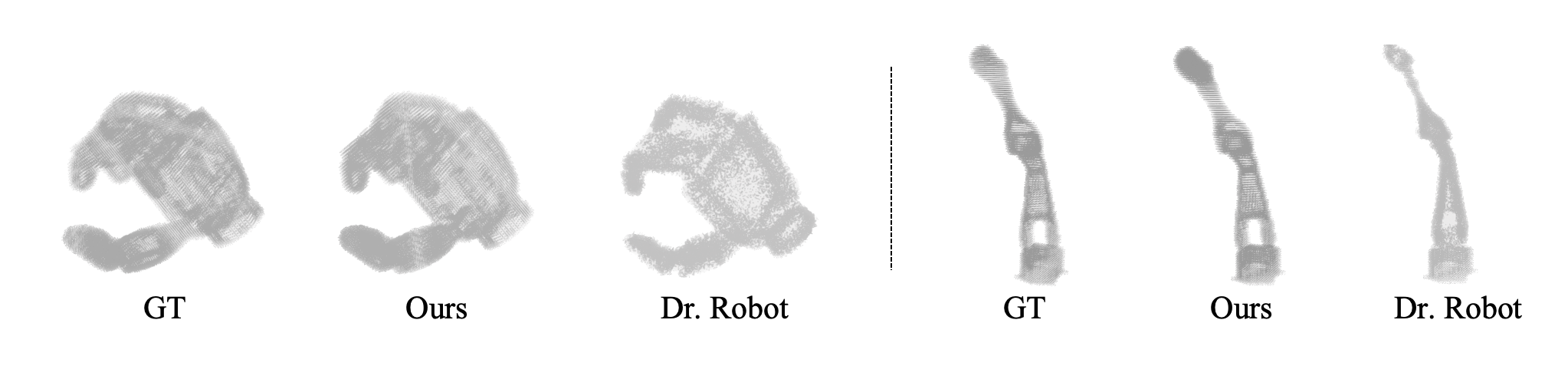}
    \caption{Qualitative comparison of geometric fidelity between WorldString and a Gaussian Splatting-based approach (\textit{Dr. Robot}) on articulated object reconstruction.}
    \label{fig:awr_vs_drrobot}
\end{figure}

\begin{table}[h]
\centering
\caption{Performance on Articulated Objects.}
\renewcommand{\arraystretch}{1.2}
\setlength{\tabcolsep}{4pt}

\resizebox{\linewidth}{!}{%
\begin{tabular}{l|llll|llll|llll|llll}
\toprule
          & \multicolumn{4}{c|}{{\underline {Robot 1 Hand}}}                                                                                        & \multicolumn{4}{c|}{{\underline {Robot 2 Arm}}}                                                                                        & \multicolumn{4}{c|}{{\underline {Furniture 21}}}                                                                                    & \multicolumn{4}{c}{{\underline {Furniture 09}}}                                                                                    \\
Type      & \multicolumn{4}{c|}{Articulated}                                                                                             & \multicolumn{4}{c|}{Articulated}                                                                                             & \multicolumn{4}{c|}{Articulated}                                                                                              & \multicolumn{4}{c}{Articulated}                                                                                              \\ \hline
Visualization
& \multicolumn{4}{c|}{\includegraphics[width=0.12\textwidth]{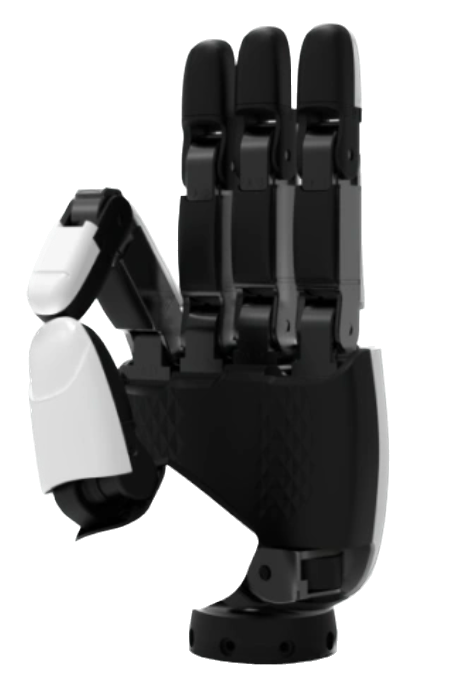}}
& \multicolumn{4}{c|}{\includegraphics[width=0.15\textwidth]{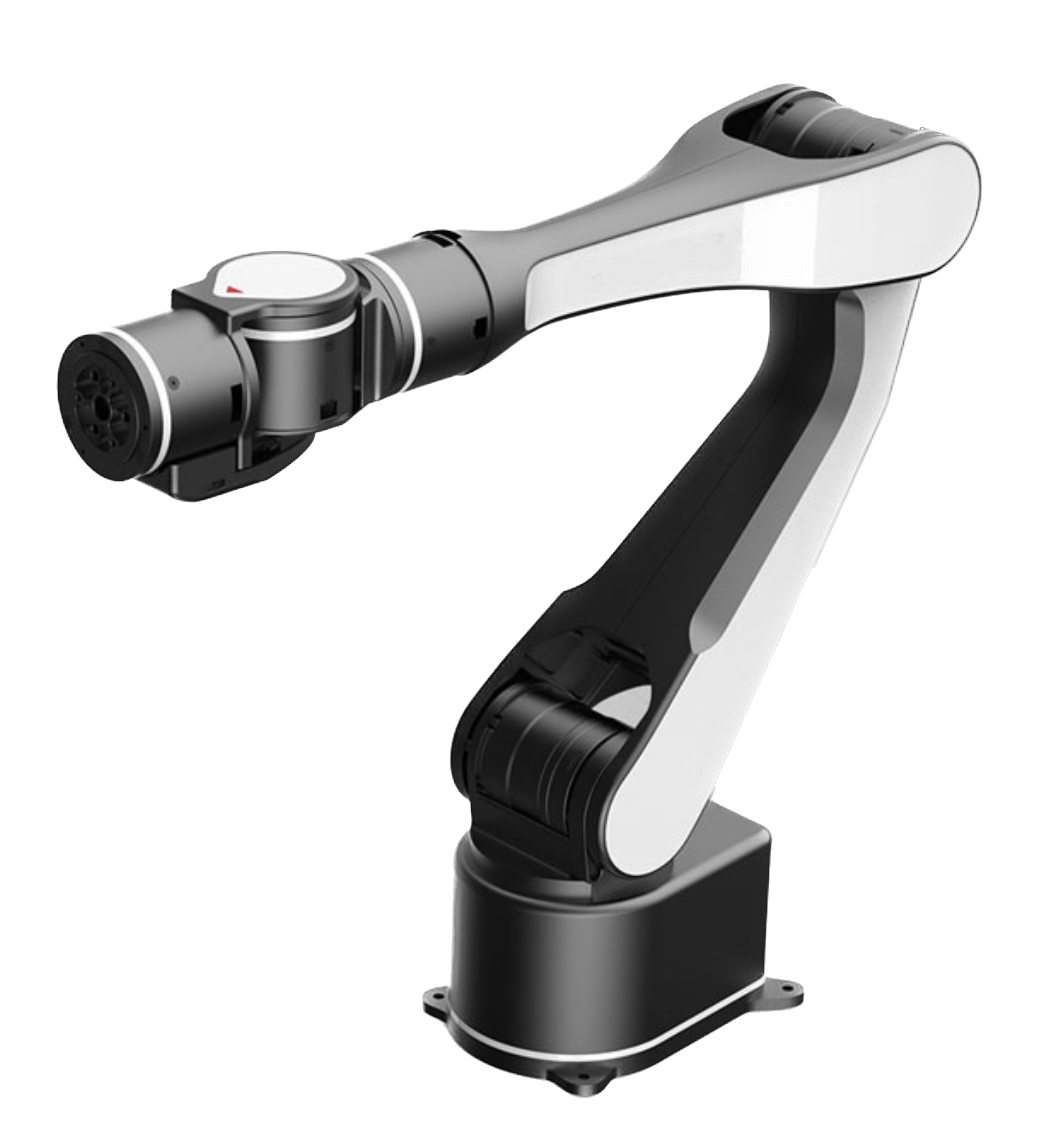}}
& \multicolumn{4}{c|}{\includegraphics[width=0.2\textwidth]{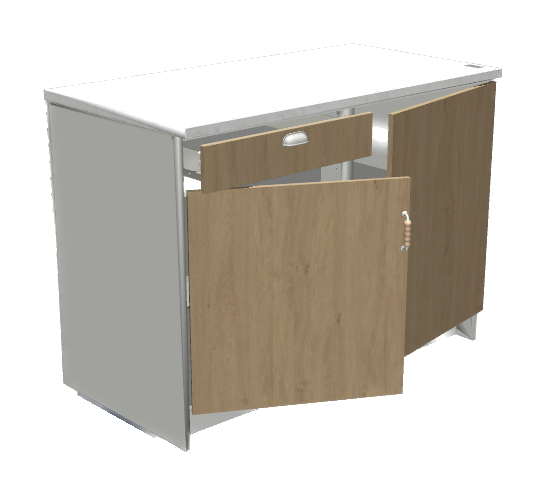}}
& \multicolumn{4}{c}{\includegraphics[width=0.15\textwidth]{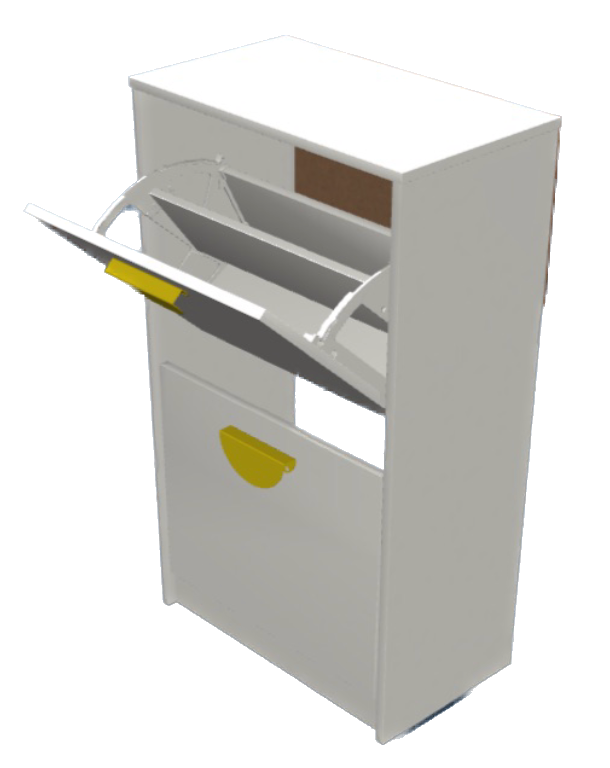}} \\
\hline
Metrics   & \multicolumn{1}{c}{{ $\mathrm{IoU}\!\uparrow$}} & \multicolumn{1}{c}{{ $F_{1}\!\uparrow$}} & \multicolumn{1}{c}{{ $\mathcal{P}\!\uparrow$}} & \multicolumn{1}{c|}{{ $\mathcal{R}\!\uparrow$}}
          & \multicolumn{1}{c}{{ $\mathrm{IoU}\!\uparrow$}} & \multicolumn{1}{c}{{ $F_{1}\!\uparrow$}} & \multicolumn{1}{c}{{ $\mathcal{P}\!\uparrow$}} & \multicolumn{1}{c|}{{ $\mathcal{R}\!\uparrow$}}
          & \multicolumn{1}{c}{{ $\mathrm{IoU}\!\uparrow$}} & \multicolumn{1}{c}{{ $F_{1}\!\uparrow$}} & \multicolumn{1}{c}{{ $\mathcal{P}\!\uparrow$}} & \multicolumn{1}{c|}{{ $\mathcal{R}\!\uparrow$}}
          & \multicolumn{1}{c}{{ $\mathrm{IoU}\!\uparrow$}} & \multicolumn{1}{c}{{ $F_{1}\!\uparrow$}} & \multicolumn{1}{c}{{ $\mathcal{P}\!\uparrow$}} & \multicolumn{1}{c}{{ $\mathcal{R}\!\uparrow$}} \\ \hline
NN        & 60.71                          & 75.39                         & 75.63                        & 75.20                         & 30.29                          & 45.52                         & 45.21                        & 45.87                         & 74.21                          & 85.16                         & 85.20                        & 85.13                         & 49.21                          & 65.17                         & 64.69                        & 65.68                        \\
Optim. NN & 73.41                          & 84.58                         & 85.36                        & 83.88                         & 47.25                          & 63.19                         & 61.94                        & 64.57                         & 31.62                          & 46.65                         & 47.68                        & 45.74                         & 38.18                          & 54.31                         & 52.61                        & 56.19                        \\
Dr. Robot   & 28.53                          & 44.31                         & 48.47                        & 40.84                        & 57.43                          & 72.94                         & 67.87                        & 78.90                         & 57.36                          & 72.90                         & 70.92                        & 75.01                         & 35.84                          & 52.76                         & 54.91                        & 50.78                        \\
\ours       & \textbf{90.28}                          & \textbf{94.89}                         & \textbf{90.87}                        & \textbf{99.28}                         & \textbf{77.00}                          & \textbf{87.01}                         & \textbf{79.55}                        & \textbf{96.01}                         & \textbf{90.17}                          & \textbf{94.83}                         & \textbf{90.49}                        & \textbf{99.61}                         & \textbf{88.98}                          & \textbf{94.17}                         & \textbf{89.51}                        & \textbf{99.35}                        \\ \bottomrule
\end{tabular}%
}
\label{table:articulated_object}
\end{table}
\noindent\textbf{Comparison with Dr. Robot.} WorldString significantly outperforms \textit{Dr. Robot} in all quantitative geometric metrics. While \textit{Dr. Robot} captures the general motion of robotic arms, its representation is composed of a collection of discrete Gaussian kernels, which leads to noisy surfaces and difficulty in representing thin, sharp mechanical structures. As shown in Fig.~\ref{fig:awr_vs_drrobot}, WorldString produces clean surfaces that precisely align with the mechanical components, whereas \textit{Dr. Robot} exhibits redundant point clusters and hollow regions within the structure. 

\subsection{Skinning-based Humans and Animals}
The quantitative results for humans and animals (Table~\ref{table:skinning_object}) further demonstrate \ours's exceptional modeling fidelity. For these categories, we specifically select keypoints that correspond to the skeletal joint positions defined by the SMPL \cite{SMPL:2015} and SMAL \cite{Zuffi:CVPR:2017} models. This deliberate alignment of input (skeletal joints) and output (shape of human or animal) spaces enables WorldString to function as a \textit{direct neural surrogate} for these classic parametric models. Our high scores across all benchmarks suggest that WorldString can effectively serve as a topology-agnostic and highly flexible alternative for complex biological skinning.
\begin{table}[h]
\centering
\caption{Performance on Skinning-based Humans and Animals.}
\vspace{-5pt}
\renewcommand{\arraystretch}{1.2}
\setlength{\tabcolsep}{4pt}

\resizebox{\linewidth}{!}{%
\begin{tabular}{l|llll|llll|llll|llll}
\toprule
          & \multicolumn{4}{c|}{{\underline {Male}}}                                                                                        & \multicolumn{4}{c|}{{\underline {Female}}}                                                                                        & \multicolumn{4}{c|}{{\underline {Horse}}}                                                                                    & \multicolumn{4}{c}{{\underline {Hippo}}}                                                                                    \\
Type      & \multicolumn{4}{c|}{Skeleton}                                                                                             & \multicolumn{4}{c|}{Skeleton}                                                                                             & \multicolumn{4}{c|}{Skeleton}                                                                                              & \multicolumn{4}{c}{Skeleton}                                                                                              \\ \hline
Visualization
& \multicolumn{4}{c|}{\includegraphics[width=0.1\textwidth]{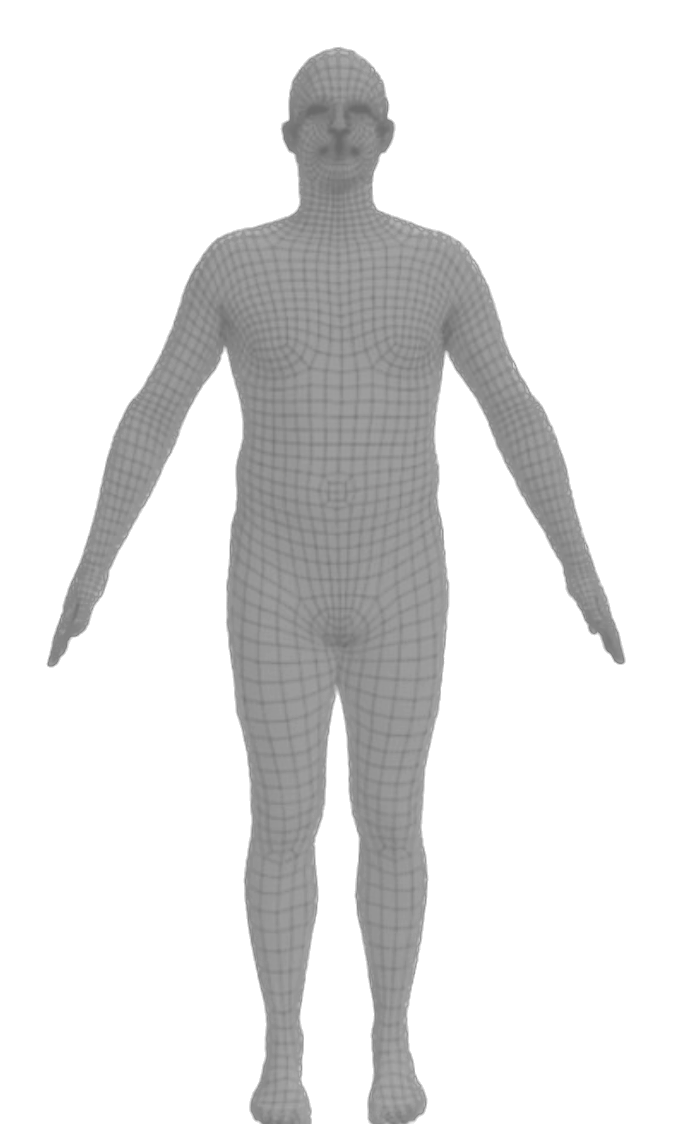}}
& \multicolumn{4}{c|}{\includegraphics[width=0.1\textwidth]{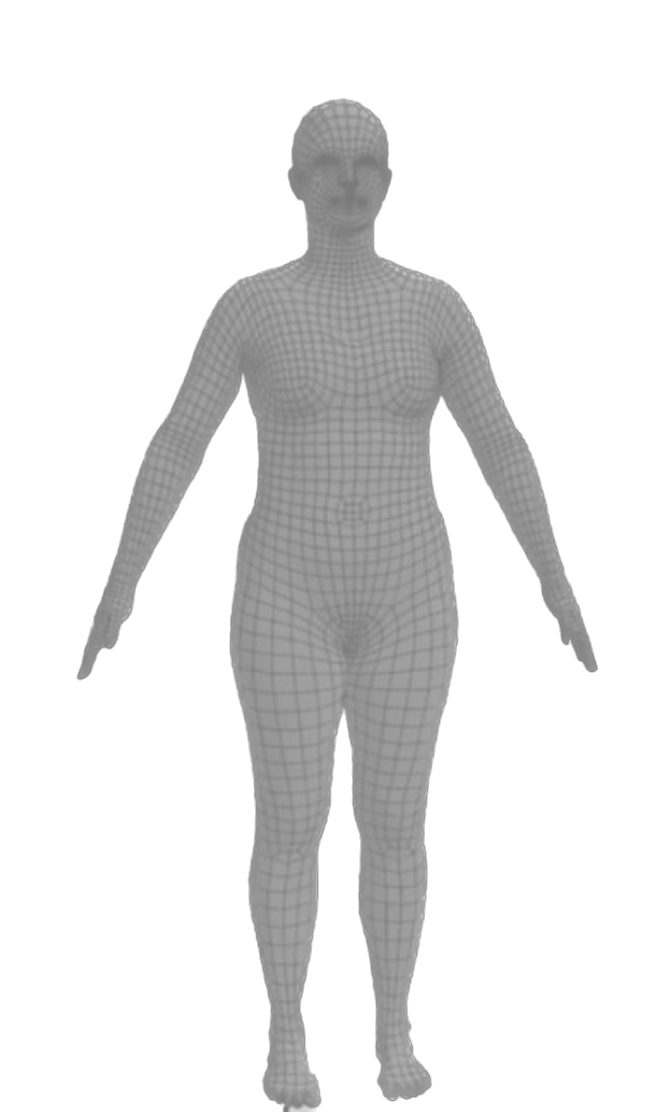}}
& \multicolumn{4}{c|}{\includegraphics[width=0.17\textwidth]{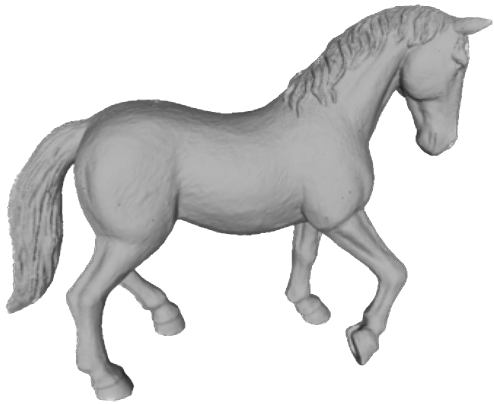}}
& \multicolumn{4}{c}{\includegraphics[width=0.25\textwidth]{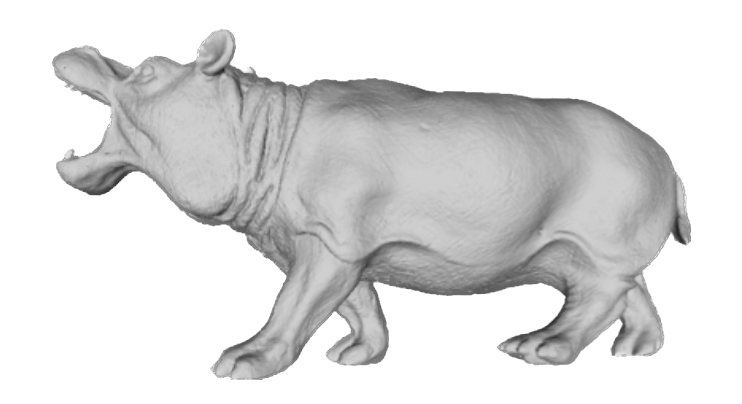}} \\
\hline
Metrics   & \multicolumn{1}{c}{{ $\mathrm{IoU}\!\uparrow$}} & \multicolumn{1}{c}{{ $F_{1}\!\uparrow$}} & \multicolumn{1}{c}{{ $\mathcal{P}\!\uparrow$}} & \multicolumn{1}{c|}{{ $\mathcal{R}\!\uparrow$}}
          & \multicolumn{1}{c}{{ $\mathrm{IoU}\!\uparrow$}} & \multicolumn{1}{c}{{ $F_{1}\!\uparrow$}} & \multicolumn{1}{c}{{ $\mathcal{P}\!\uparrow$}} & \multicolumn{1}{c|}{{ $\mathcal{R}\!\uparrow$}}
          & \multicolumn{1}{c}{{ $\mathrm{IoU}\!\uparrow$}} & \multicolumn{1}{c}{{ $F_{1}\!\uparrow$}} & \multicolumn{1}{c}{{ $\mathcal{P}\!\uparrow$}} & \multicolumn{1}{c|}{{ $\mathcal{R}\!\uparrow$}}
          & \multicolumn{1}{c}{{ $\mathrm{IoU}\!\uparrow$}} & \multicolumn{1}{c}{{ $F_{1}\!\uparrow$}} & \multicolumn{1}{c}{{ $\mathcal{P}\!\uparrow$}} & \multicolumn{1}{c}{{ $\mathcal{R}\!\uparrow$}} \\ \hline
NN        & 40.31                          & 57.22                         & 57.17                        & 57.29                         & 43.61                          & 60.49                         & 60.23                        & 60.77                         & 35.52                          & 51.55                         & 51.68                        & 51.43                         & 41.21                          & 57.38                         & 57.71                        & 57.07                        \\
Optim. NN & 55.64                          & 71.29                         & 72.34                        & 70.29                         & 60.35                          & 75.13                         & 75.55                        & 74.72                         & 74.85                          & 85.58                         & 86.15                        & 85.02                         & 79.33                          & 88.43                         & 89.40                        & 87.50                        \\
NSDP & 67.41                          & 80.46                         & 87.02                        & 75.03                         & 70.13                          & 82.38                         & \textbf{93.89}                        & 73.45                         & 76.25                          & 86.51                         & 81.69                        & 91.95                      & 86.82                          & 92.91                         & 95.52                        & 90.46                        \\
\ours       & \textbf{83.47}                          & \textbf{90.99}                         & \textbf{88.37}                        & \textbf{93.78}                         & \textbf{87.83}                          & \textbf{93.52}                         & 91.29                        & \textbf{95.86}                         & \textbf{90.54}                          & \textbf{95.04}                         & \textbf{93.70}                        & \textbf{96.41}                         & \textbf{92.40}                          & \textbf{96.05}                         & \textbf{95.96}                        & \textbf{96.15}                        \\ \bottomrule
\end{tabular}%
}
\label{table:skinning_object}
\end{table}

\begin{figure}
    \centering
    \includegraphics[width=0.9\linewidth]{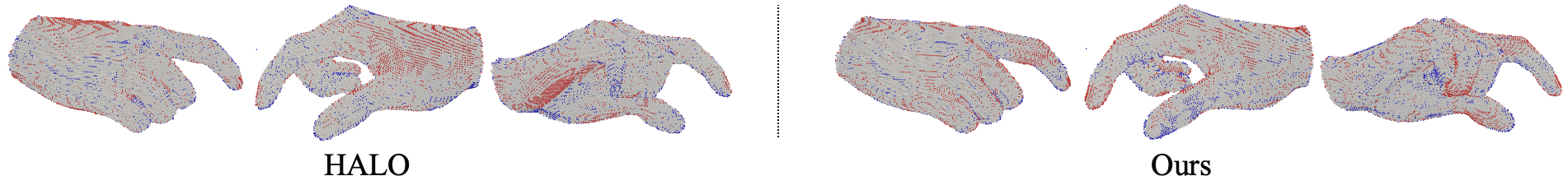}
    \caption{Qualitative error-map comparison between HALO and \ours\ on hand reconstruction. \textbf{Gray:} correct occupancy prediction; \textbf{red:} false positives; \textbf{blue:} false negatives.}
    \label{fig:awr_vs_halo}
\end{figure}

\begin{wraptable}{r}{0.33\textwidth}
\centering
\vspace{-10pt}
\caption{Performance on Hand.}
\renewcommand{\arraystretch}{1.2}
\setlength{\tabcolsep}{4pt}

\resizebox{1.\linewidth}{!}{%
\begin{tabular}{l|llll}
\toprule
          & \multicolumn{4}{c}{{\underline {Hand}}}\\
Type      & \multicolumn{4}{c}{Skeleton}\\ \hline
Visualization
& \multicolumn{4}{c}{\includegraphics[width=0.1\textwidth]{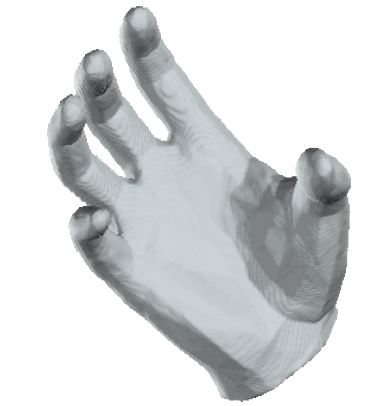}}\\
\hline
Metrics   & \multicolumn{1}{c}{{ $\mathrm{IoU}\!\uparrow$}} & \multicolumn{1}{c}{{ $F_{1}\!\uparrow$}} & \multicolumn{1}{c}{{ $\mathcal{P}\!\uparrow$}} & \multicolumn{1}{c}{{ $\mathcal{R}\!\uparrow$}}\\ \hline
HALO        & \textbf{96.62}                          & \textbf{98.28}                         & \textbf{98.15}                        & 98.40    \\
WorldString        & 96.24                          & 98.08                         & 97.43                        & \textbf{98.74}
\\ \bottomrule
\end{tabular}%
}
\label{table:halo_object}
\end{wraptable}
\noindent\textbf{Comparison with NSDP and HALO.} 
NSDP~\cite{tang2022neural} predict mesh deformations from sparse user ``handles'' which reduce to \emph{part of surface shape and position} at limb tips and the head. WorldString achieves higher volumetric scores than NSDP across human and animal categories, indicating that a single keypoint-conditioned occupancy decoder transfers more readily across bipeds and quadrupeds than deformation priors centered on handle-driven quadruped setups.
HALO~\cite{karunratanakul2021halo} use 3D joint locations drive a skeleton-conditioned neural occupancy field for the posed hand.
Table~\ref{table:halo_object} shows that WorldString matches HALO within a narrow margin on IoU, $F_1$, precision, and recall---both models attain excellent hand occupancy fidelity under comparable supervision.
Fig.~\ref{fig:awr_vs_halo} complements Table~\ref{table:halo_object} with a qualitative error-map visualization on matched hand poses. The remaining red and blue points for both method are sparse and concentrated in fine-scale regions.
The practical difference is therefore \emph{generality}: HALO is restricted to human hands, whereas \ours\ applies the same architecture to all kinds of objects and deformation types.

\subsection{Real World Soft Bodies}
WorldString demonstrates robust performance in modeling high DoF non-linear manifolds. We provide a detailed description in the Appendix for real world data acquisition. In Table~\ref{table:soft_object}, we observe a nuanced result for the \textit{Rope} category: The Optim. NN baseline achieves competitive scores in certain metrics. This is attributed to the relatively low-dim deformation space of a short rope, where the combination of state retrieval and IDW-based local refinement can accurately approximate simple bending motions. However, for more complex soft interactions where deformation is non-homogeneous, \ours’s learned implicit representation proves more capable of preserving volume and surface consistency.
\noindent

\begin{table}[h]
\centering
\caption{Performance on Soft Objects.}
\renewcommand{\arraystretch}{1.2}
\setlength{\tabcolsep}{4pt}

\resizebox{\linewidth}{!}{%
\begin{tabular}{l|llll|llll|llll}
\toprule
          & \multicolumn{4}{c|}{{\underline {Doll}}}                                                                                        & \multicolumn{4}{c|}{{\underline {Cloth}}}                                                                                        & \multicolumn{4}{c}{{\underline {Rope}}}                                                                                    \\
Type      & \multicolumn{4}{c|}{Deformable}                                                                                             & \multicolumn{4}{c|}{Deformable}                                                                                             & \multicolumn{4}{c}{Deformable}                                                                                             \\ \hline
Visualization
& \multicolumn{4}{c|}{\includegraphics[width=0.14\textwidth]{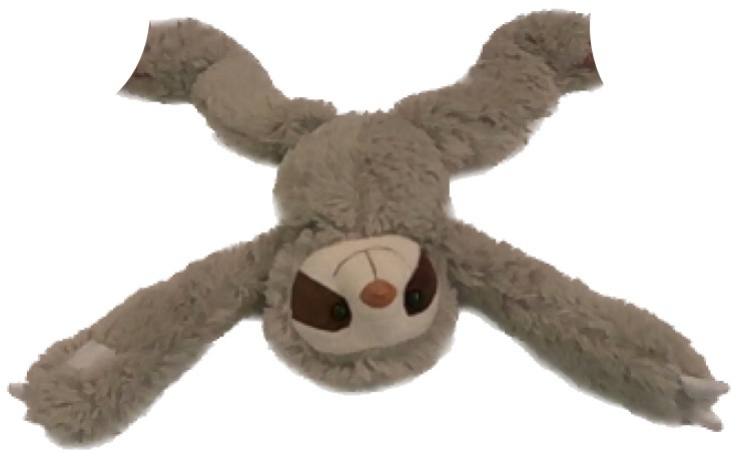}}
& \multicolumn{4}{c|}{\includegraphics[width=0.14\textwidth]{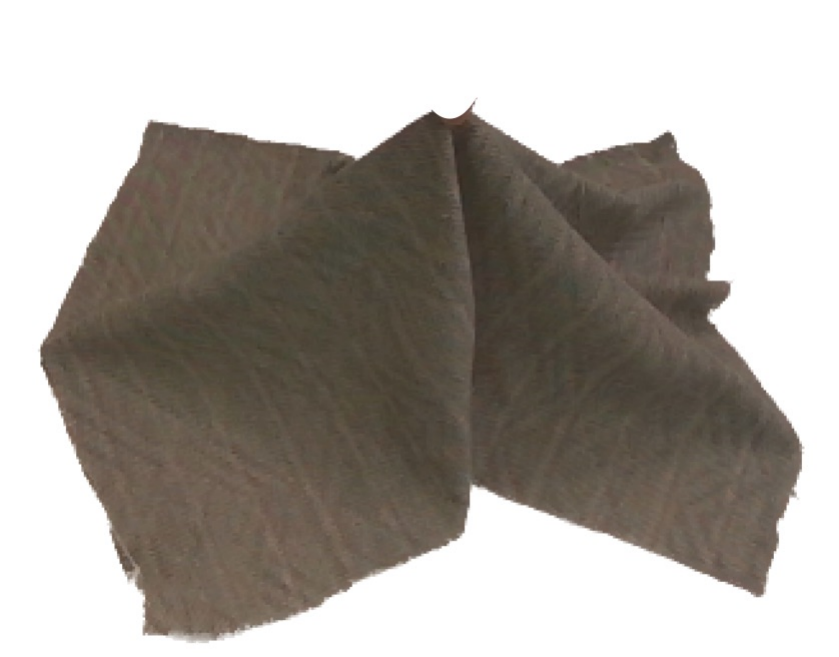}}
& \multicolumn{4}{c}{\includegraphics[width=0.15\textwidth]{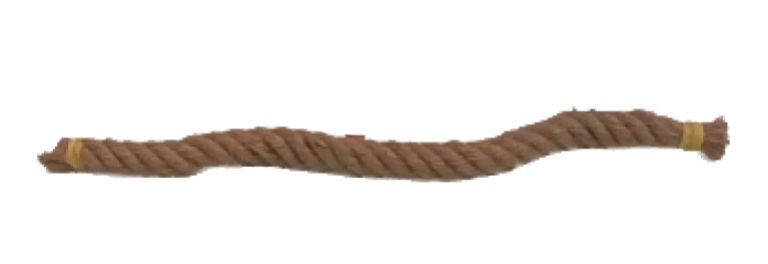}} \\
\hline
Metrics   & \multicolumn{1}{c}{{ $\mathrm{IoU}\!\uparrow$}} & \multicolumn{1}{c}{{ $F_{1}\!\uparrow$}} & \multicolumn{1}{c}{{ $\mathcal{P}\!\uparrow$}} & \multicolumn{1}{c|}{{ $\mathcal{R}\!\uparrow$}}
          & \multicolumn{1}{c}{{ $\mathrm{IoU}\!\uparrow$}} & \multicolumn{1}{c}{{ $F_{1}\!\uparrow$}} & \multicolumn{1}{c}{{ $\mathcal{P}\!\uparrow$}} & \multicolumn{1}{c|}{{ $\mathcal{R}\!\uparrow$}}
          & \multicolumn{1}{c}{{ $\mathrm{IoU}\!\uparrow$}} & \multicolumn{1}{c}{{ $F_{1}\!\uparrow$}} & \multicolumn{1}{c}{{ $\mathcal{P}\!\uparrow$}} & \multicolumn{1}{c}{{ $\mathcal{R}\!\uparrow$}} \\ \hline
NN        & 44.90                          & 59.78                         & 60.00                        & 59.76                         & 46.80                          & 61.58                         & 62.42                        & 61.59                         & 61.27                          & 74.55                         & 74.22                        & 74.89                        \\
Optim. NN & 61.58                          & 75.22                         & 74.13                        & 76.68                         & 41.91                          & 56.47                         & 58.50                        & 54.71                         & \textbf{79.64}                          & \textbf{88.65}                         & \textbf{87.80}                        & 89.55                        \\
\ours       & \textbf{82.80}                          & \textbf{90.59}                         & \textbf{84.92}                        & \textbf{97.07}                         & \textbf{68.68}                          & \textbf{81.43}                         & \textbf{71.20}                        & \textbf{95.09}                         & 78.34                          & 87.85                         & 81.94                        & \textbf{94.68}                        \\ \bottomrule
\end{tabular}%
}
\label{table:soft_object}
\end{table}

\vspace{-10pt}
\subsection{Effectiveness and Robustness on Noisy Sensor Observations}
\label{sec:real_world_analysis}

\begin{wrapfigure}{r}{0.48\textwidth}
  \centering
  \vspace{-20pt}
  \includegraphics[width=0.5\textwidth]{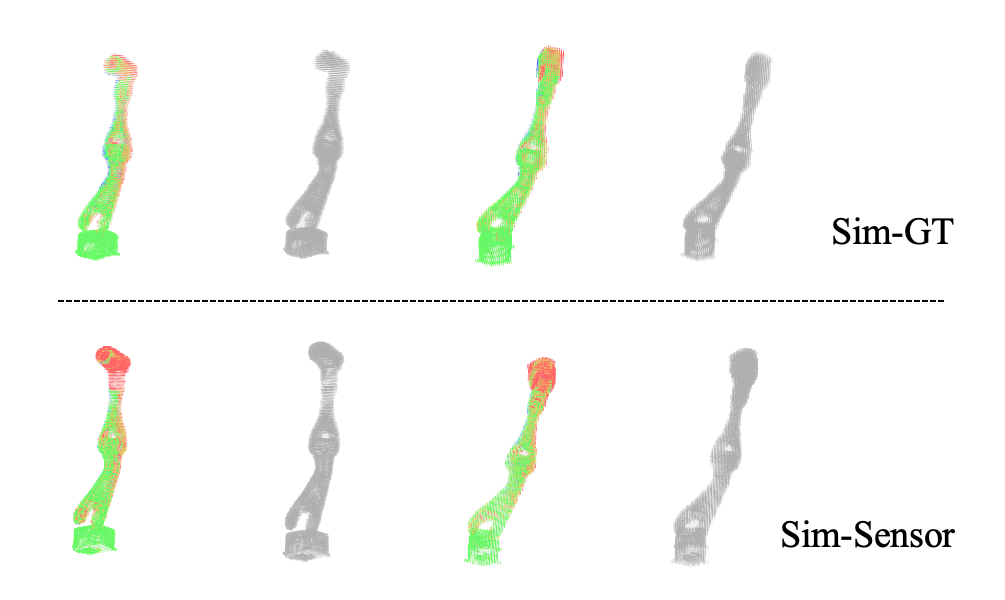}
  \vspace{-20pt}
  \caption{Qualitative visualization of structural completion in our gap study.}
  \label{fig:gap_study}
  \vspace{-27pt}
\end{wrapfigure}

A critical concern is whether the real world data acquisition pipeline introduces significant noise or systematic bias that hinders the model's learning. If the WorldString cannot handle such inherent sensor imperfections, scaling up to real-world objects would be completely infeasible. To address this, we evaluate \ours's robustness through a progressive analysis: from an in-silico gap study to real-world observations.

\begin{wraptable}{r}{0.36\textwidth}
  \centering
  \caption{Quantitative \textit{in-silico} gap study evaluating the impact of sensor noise.}
  \label{tab:gap_study}
  \renewcommand{\arraystretch}{1.2}
  \setlength{\tabcolsep}{4pt}
  \resizebox{\linewidth}{!}{%
  \begin{tabular}{l|llll}
  \toprule
            & \multicolumn{4}{c}{{\underline {Robot Arm}}} \\
  Type      & \multicolumn{4}{c}{Articulated} \\ \hline
  Metrics   & \multicolumn{1}{c}{{ $\mathrm{IoU}\!\uparrow$}} & \multicolumn{1}{c}{{ $F_{1}\!\uparrow$}} & \multicolumn{1}{c}{{ $\mathcal{P}\!\uparrow$}} & \multicolumn{1}{c}{{ $\mathcal{R}\!\uparrow$}} \\ \hline
  Sim-Sensor        & 60.20                          & 75.15                          & 61.82                        & 95.81                          \\
  Sim-GT & 77.00                          & 87.01                          & 79.55                        & 96.01                          \\ \bottomrule
  \end{tabular}%
  }
\vspace{-10pt}
\end{wraptable}

\noindent\textbf{Quantifying the Sensor Gap and Structural Completion.} Since obtaining perfect ground-truth (GT) geometry in real-world settings is physically impossible, we first conduct a validation study. We replicate the multi-view RGB-D capture pipeline within a physics simulator to generate "Sim-Sensor" data, which is then compared against the simulator's native "Sim-GT" geometry using a robot arm. As presented in Table~\ref{tab:gap_study}, we evaluate the quantitative performance of WorldString trained on these two data sources. Crucially, while the sensor-fusion process inevitably introduces discretization artifacts, the $F_1$ score does not suffer a catastrophic collapse. This indicates that the model successfully avoids representation collapse and still captures the essential actionable manifold despite the degraded input. 

Furthermore, our gap study yields an important qualitative finding regarding \textit{structural completion}. As shown in the visualization of the robot arm (Fig.~\ref{fig:gap_study}), the simulated cameras fail to capture certain parts of the geometry due to self-occlusion. However, \ours's prediction successfully completes these missing structures. This demonstrates that training on noisy sensor data actually triggers the model's emergent capability to recover unobserved geometries.

\begin{figure}[t]
    \centering
    \includegraphics[width=0.75\linewidth]{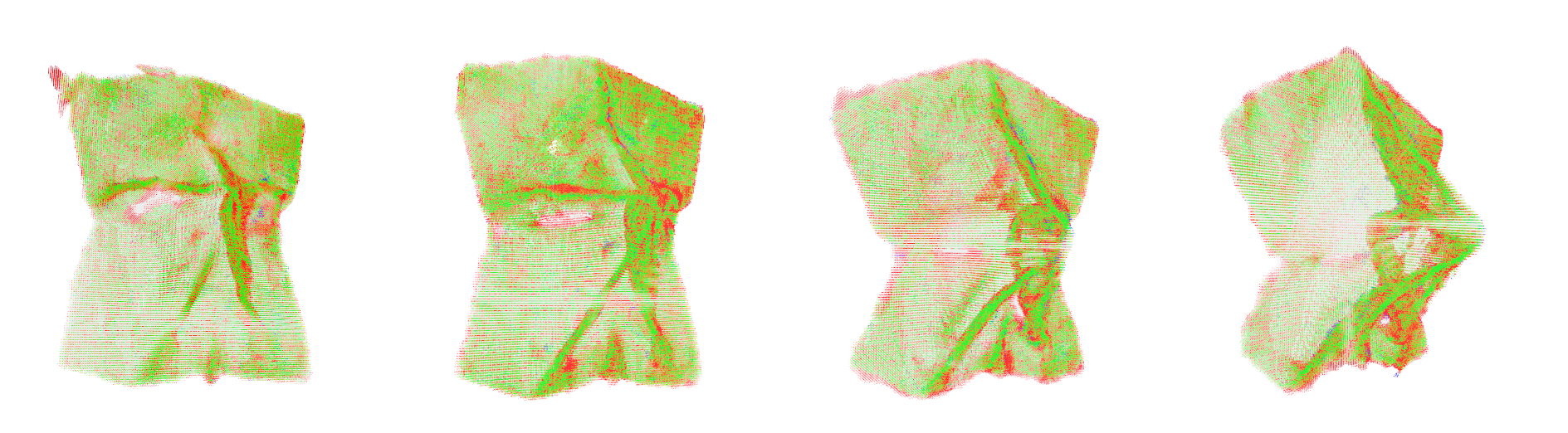}
    \caption{Visualization of WorldString's robust predictions on real-world cloth sequences. Green, red, and blue denote true positives, false positives, and false negatives, respectively. The visualization shows that some of the false positives points are an auto completion of the missing parts.}
    \label{fig:rubust}
    \vspace{-10pt}
\end{figure}

\noindent\textbf{Robustness and Material Completion on Real Data.} 
Fig.~\ref{fig:rubust} presents an error map analysis on real-world cloth sequences.
The almost complete absence of blue points confirms that the model robustly remembers the full object structure without omissions. More interestingly, we observe a second, distinct type of completion phenomenon. A significant amount of red points are scattered uniformly across the fabric regions. Because real-world RGB-D sensors inherently produce sparse point clouds, the captured "ground truth" used for evaluation often contains artificial "holes" on what is actually a dense material. 

The presence of these red predictions indicates that WorldString recognizes that the cloth is a continuous, solid fabric and actively fills in the missing sensory gaps, reconstructing a dense manifold that reflects the physical reality. This dual capability—structural completion for occlusions (as seen in the robot arm) and material completion for sensory sparsity (as seen in the cloth)—proves that WorldString leverages its representation to robustly infer physical reality.

\subsection{Interpretability of 3D Shape Tokens}
\label{sec:interpretability}
\noindent\textbf{Visualization Mechanism.} 
The core of our interpretability analysis lies in attributing each predicted occupancy point to its most influential query tokens. During inference, for any spatial query point $\mathbf{s}$, we identify the top-$5$ query tokens that assign the highest attention weights to $\mathbf{s}$ in the cross-attention layer. To visualize this relationship, we assign a unique, fixed color to each query token in the canonical space. The final color of a predicted 3D point is computed as a weighted sum of the colors of these top-$5$ tokens, where the weights are derived from their respective normalized attention scores.

\begin{figure}
    \centering
    \includegraphics[width=0.9\linewidth]{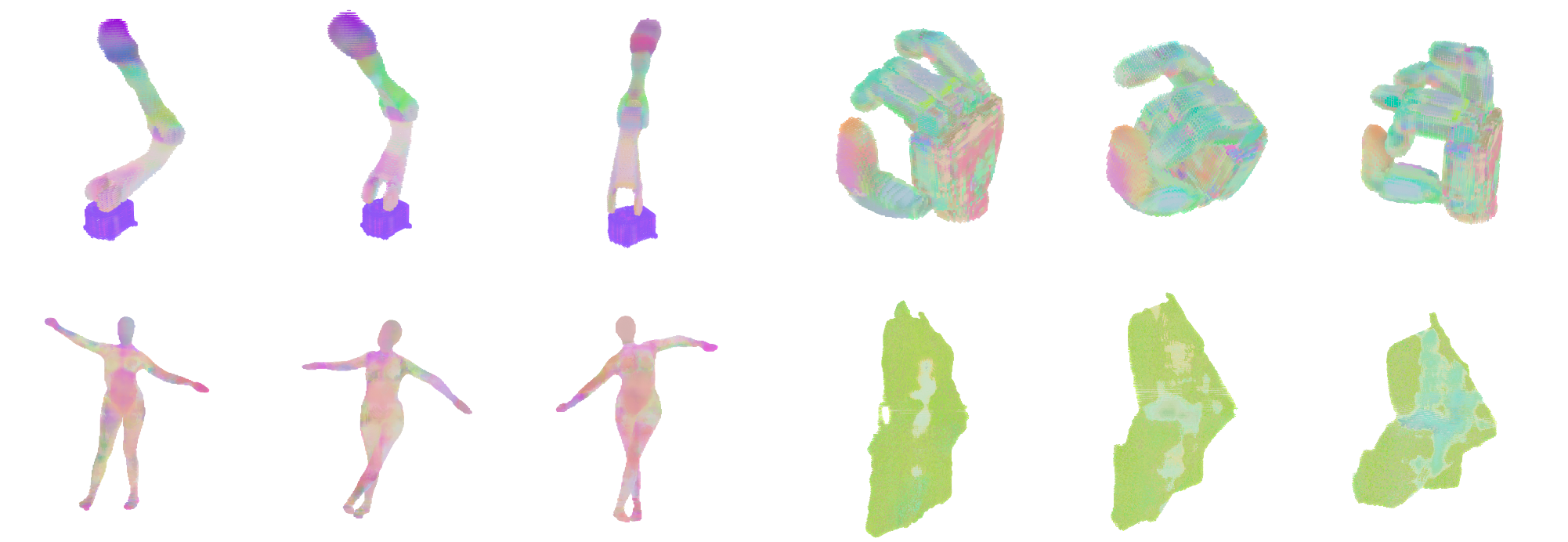}
    \caption{\textbf{Interpretability of WorldString's latent representation.} Each predicted spatial point is colored based on a weighted sum of its top5-attending query tokens.}
    \label{fig:interpretability}
\end{figure}

\noindent\textbf{Pose-Invariant Part Specialization.} 
As shown in Fig.~\ref{fig:interpretability}, this visualization reveals a striking emergent property: \textbf{semantic consistency across varied poses}. Despite significant articulations, specific physical parts of the object consistently exhibit the similar color signatures. For instance, in the \textit{Xhand} sequences, the outer surface of the thumb consistently maintains a pink hue regardless of the gesture. Similarly, in the \textit{Human Body} reconstructions, both hands are consistently attributed a purple color signature across a wide range of diverse and complex postures.

\begin{wraptable}{r}{0.5\textwidth}
\centering
\vspace{-20pt}
\caption{Ablation Study on Robot Arm. We investigate the impact of attention layers ($L$), hidden dimension ($D$), spatial resolution ($R$), and keypoint density ($K$).}
\label{table:ablation_study}
\renewcommand{\arraystretch}{1.1}
\setlength{\tabcolsep}{5pt}
\resizebox{\linewidth}{!}{%
\begin{tabular}{cccc|cccc}
\toprule
\multicolumn{4}{c|}{Model Parameters} & \multicolumn{4}{c}{Metrics} \\
$L$ & $D$ & $R$ & $K$ & $\mathrm{IoU}\!\uparrow$ & $F_{1}\!\uparrow$ & $\mathcal{P}\!\uparrow$ & $\mathcal{R}\!\uparrow$ \\ \midrule
2 & 128 & 512 & 3  & 77.00 & 87.01 & 79.55 & 96.01 \\
2 & 128 & 512 & 15 & \textbf{83.51} & \textbf{91.02} & 84.79 & \textbf{99.46} \\ \hline
1 & 128 & 512 & 3  & 71.16 & 83.15 & 72.61 & 97.28 \\
1 & 192 & 512 & 3  & 68.78 & 81.50 & 72.34 & 93.32 \\
2 & 64  & 512 & 3  & 72.54 & 84.09 & 74.07 & 97.24 \\
2 & 192 & 512 & 3  & 72.86 & 84.30 & 74.85 & 96.49 \\
3 & 64  & 512 & 3  & 73.03 & 84.41 & 74.67 & 97.09 \\
3 & 128 & 512 & 3  & 70.95 & 83.01 & 76.02 & 91.41 \\ \hline
2 & 128 & 768 & 3  & 74.42 & 85.33 & 75.97 & 97.32 \\
2 & 128 & 256 & 3  & 82.37 & 90.33 & \textbf{85.31} & 95.98 \\ \bottomrule
\end{tabular}%
}
\vspace{-30pt}
\end{wraptable}

\noindent\textbf{Discussion on Structural Anchoring.} 
These observations provide strong evidence that the \ours model does not treat the object as a holistic, unstructured volume. Instead, each query token learns to specialize in representing a relatively fixed, localized segment of the object's canonical geometry. This emergent part-based decomposition is a direct result of the cross-attention mechanism between shape tokens and input keypoints. By attending to the structural keypoints, the latent queries are effectively "anchored" to the underlying physical manifold. This experiment confirms that our keypoint-driven input provides a robust structural prior, enabling the model to learn a disentangled and interpretable representation of complex actionable objects.
\subsection{Ablation study}
\label{sec:ablation}

We conduct ablation experiments to analyze the impact of keypoint density, voxel resolution, and network capacity on \ours's performance. The quantitative results are summarized in Table~\ref{table:ablation_study}.

\noindent\textbf{Keypoint Density.} 
We observe that increasing the number of keypoints per component (e.g., to 15 points) improves reconstruction accuracy. While theoretically three non-collinear points are sufficient to determine the 6-DoF pose of a rigid part, denser keypoints provide redundant but crucial geometric structural information. This extra supervision makes it easier for the model to "anchor" the shape tokens to the underlying manifold, facilitating the learning of intricate local geometries.

\noindent\textbf{Voxel Resolution.} 
The results indicate that higher spatial resolutions increase the complexity of the occupancy learning task. We observe a slight performance degradation as the voxel resolution increases; however, this decline is marginal. This suggests that while finer grids impose stricter requirements on the model's boundary-fitting capability, \ours maintains robust convergence across a reasonable range of resolutions.

\noindent\textbf{Network Capacity.} 
Interestingly, the ablation study reveals that merely increasing the network's overall parameter count or architectural depth does not monotonically yield better results for specific object categories. For instance, as detailed in Table \ref{table:ablation_study}, elevating the hidden dimension ($D$) from 128 to 192, or increasing the number of attention layers ($L$) from 2 to 3, actually degrades overall performance across key metrics like Intersection over Union (IoU) and $F_1$ scores. This implies that for a given actionable manifold, there exists an optimal capacity threshold. Pushing the model beyond this limit introduces unnecessary complexity, which likely leads to diminishing returns or subtle overfitting---where the network begins to memorize specific training configurations rather than learning generalizable, robust geometric features. Consequently, our current baseline architecture strikes a highly favorable balance; it secures the necessary representation power to accurately model complex object interactions while preserving the computational efficiency required for practical deployment.
\section{Conclusion}
In this paper, we introduced WorldString, a unified, keypoint-driven actionable object representation. By mathematically demonstrating that classical kinematics (FK), linear blend skinning (LBS), and soft-body Jacobians can all be relaxed into a unified residual attention mechanism, we bridged the gap between rigorous physical priors and flexible neural implicit fields. Extensive experiments demonstrate that WorldString successfully models the intricate deformation manifolds of all kinds of objects under a single, topology-agnostic transformer architecture. Furthermore, our model exhibits remarkable robustness against real-world sensor noise and demonstrates emergent capabilities in structural completion and interpretable part-specialization.

\clearpage

\bibliographystyle{abbrvnat}
\nobibliography*
\bibliography{googledeepmind-test}

\end{document}